%% file: main.tex
\documentclass{article}
\usepackage{iclr2026_conference_arxiv,times}
\usepackage[utf8]{inputenc}
\usepackage[T1]{fontenc}
\usepackage{hyperref}
\usepackage{url}
\usepackage{amsmath,amssymb}
\usepackage{graphicx}
\usepackage{booktabs}
\usepackage{multirow}
\usepackage{subcaption}
\usepackage{xcolor}
\usepackage{ifthen}
\usepackage{enumitem}
\usepackage{wrapfig}
\usepackage[capitalize,noabbrev]{cleveref}
\usepackage{tikz}
\usetikzlibrary{fadings,shapes}
\usetikzlibrary{shadows}
\usepackage{amsthm}

\usepackage{pgfplots}
\usepackage{makecell}
\pgfplotsset{compat=1.16}
\usepgfplotslibrary{groupplots}
\input{figures/fig_algo_preamble}

\usepackage[table]{xcolor}
\definecolor{tableShade}{RGB}{241,247,237}

\title{Atlases Are Already Inside: \\Recovering Population Templates from  Pretrained Diffusion Models}

\author{%
  \begin{minipage}{\dimexpr\textwidth-2\tabcolsep\relax}\centering\normalfont
  ~\\[9pt]
  {\bf Jian Shi}$^{1}$ \quad {\bf John Femiani}$^{2}$ \quad  {\bf Peter Wonka}$^{1}$ \\[9pt]
  $^{1}$KAUST \qquad $^{2}$Miami University \\[3pt]
  \end{minipage}%
}

\iclrarxivcopy

\newcommand{\eg}{\emph{e.g.}}

\begin{document}

\maketitle

\begin{figure}[h]
    \centering
    \input{figures/teaser/fig_teaser_fancy}
    \caption{
    \textbf{Recovery of a population's central anatomy from diffusion models}, across different modalities, including 3D brain MRI, 2D chest X-rays, and 2D faces. Generators are trained to synthesize individuals (bottom). Our method extracts its intrinsic atlas from each generator (top).
    }
    \label{fig:teaser}
\end{figure}
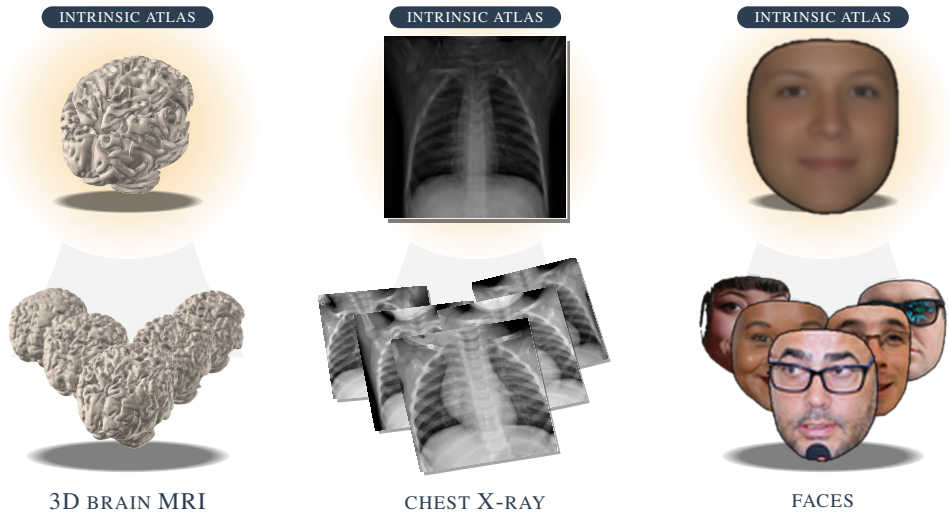

\begin{abstract}
We present a new inference-time sampler for diffusion models that gives a pretrained model a capability it was never trained for: constructing the atlas of the population it synthesizes. The sampler converges from every random seed to the population's central anatomy, which we call the \emph{intrinsic atlas}. The advantage is threefold. (1) It requires no retraining. A diffusion model that has already learned a coherent population, including the released ones, yields its atlas in a single inference pass without involving deformable registration. (2) It applies to multiple domains, such as brain MRI, chest X-ray, faces, and 3D shapes. (3) It extends to subpopulations. One age-conditioned model gives an atlas at any age in its training range, and the resulting family reproduces the CSF expansion of healthy aging. Evaluated as a registration target, the intrinsic atlas is best or second-best on every dataset against classical and learned templates, and the most central template on held-out brain MRI cohorts. Atlas construction can be reframed as a byproduct of generative modeling: a diffusion model is a learned representation of population structure, and the atlas is what it already contains.
\end{abstract}

\section{Introduction}
\label{sec:intro}

Every coherent population has a center, where every member inside it is a smooth deformation of one shared structure: the brain that all brains are variations of, the face behind all faces, the letter beneath every typeface.
\citet{grenander1998computational} formalized this center as a template, in which a population is one template together with the transformations that carry it to each member. This template is also known as an atlas in medical imaging.
In practice, fields that need an atlas build it laboriously by spatially aligning thousands of individuals into a common frame under a chosen deformation model~\citep{Klein2009,xu2019deepatlas,ding2022aladdin,ranem2024continual}.
Learning-based methods amortize alignment into a network~\citep{balakrishnan2019voxelmorph,dey2021generative,Dalca2019learning,ImplicitAtlas2022,abulnaga2025multimorph}, but a deformation model remains what defines the atlas.
In contrast, we show that the atlas can be obtained by a new sampling strategy on diffusion models~\citep{ho2020denoising,song2021scorebased} trained for synthesis alone.
This sampler converges from every random start to one sharp image (cross-seed SSIM $\ge 0.99$), on brains, faces, and 3D shapes alike. We refer to this behavior as \textit{canonical convergence}.

Our hypothesis behind it comes from a correspondence between the diffusion model and pattern theory. Pattern theory, as used in computational anatomy, decomposes a population into a template and the transformations that deform it to each member~\citep{grenander1970unified,grenander1998computational}. The reverse process of a diffusion model has the same two terms. A learned denoiser moves the state toward the posterior mean of the population given the current noisy state, and an injected noise decides where within the population the trajectory goes.
Prior work~\citep{meng2021sdedit,huberman2024edit} exploits the second term by editing the injected noise to edit samples, showing that the noise carries individual variation within the population.
Our hypothesis concerns the complementary term. We posit that the denoiser carries what the population shares, namely, the template.
If so, then for a coherent population, whose members share a single structure and differ only by continuous variation, the denoiser alone should trace out the population's central instance, which we call the \textit{intrinsic atlas} in this work. The hypothesis makes three commitments, and each is testable. First, canonical convergence must be a property of the sampler, not of one network. Second, it must serve as a valid registration target. Third, it must report the structure of the population it was trained on, returning one image where the population has one template, one image per template where it has several, and no stable image where it has none. The rest of this paper tests all three.

The first commitment holds for every population that shares a central anatomy, medical and non-medical, in 3D and 2D, with a mechanism investigation in~\cref{sec:mechanism}. For the second, we evaluate the intrinsic atlas as a registration target under a standard registration protocol, first on T1-weighted brain MRI, where atlas construction and its evaluation are most mature, then on faces, chest X-ray, and 3D shapes. With no registration, fine-tuning, or atlas objective, it is best or second-best on every dataset against classical and learned templates, and the most central and most regular template on held-out brain cohorts (\cref{sec:results}). The template also moves with the generator's conditioning, where an age-conditioned model returns a family of age-specific brain atlases on demand that reproduces the CSF expansion of healthy aging (\cref{sec:age}). For the third, populations with several templates or none resolve into multiple templates or into a blurry image (\cref{sec:failures}).
Our contributions are as follows:
\begin{itemize}[leftmargin=1.4em,itemsep=1pt,topsep=2pt]
    \item \textbf{Canonical convergence by design.} A deterministic variant of sampling turns any diffusion model of a coherent population into an atlas constructor. We explain the convergence as a contraction of the noise-free operator, and verify it with controlled experiments.
    \item \textbf{Cross-domain registration benchmark.} Under a fixed registration protocol per domain, the intrinsic atlas is best or second-best across multiple domains, such as brain MRI, faces, chest X-ray, and 3D shapes. Our method is the only method that spans them all.
    \item \textbf{Conditioning and population structure.} The atlas inherits the generator's conditioning, giving an age-specific atlas family on demand. Where a population has several centers or none, the same sampler makes that visible.
\end{itemize}

\section{Related Work}
\label{sec:related}

\subsection{Atlas and template construction}

\paragraph{Atlases in the medical field.}
Classical atlas construction alternates registration and averaging under an explicit deformation model, \eg\ \textit{LDDMM}~\citep{Beg2005} and unbiased-template diffeomorphic construction~\citep{fonov2011unbiased,Klein2009,tustison_antsx_2021}. Learning-based methods amortize this optimization into a network. \textit{VoxelMorph}~\citep{balakrishnan2019voxelmorph} and follow-ups learn pairwise or groupwise registration. \textit{Atlas-GAN}~\citep{dey2021generative} and conditional deformable-template networks~\citep{Dalca2019learning,AtlasMorph2025} learn a network whose output is the atlas, trained under a template-construction or shape-reconstruction loss. \textit{ImplicitAtlas}~\citep{ImplicitAtlas2022} represents the template implicitly. \textit{MultiMorph}~\citep{abulnaga2025multimorph} trains a feedforward model that produces a population-specific atlas in one forward pass, reporting a $100\times$ speedup over iterative construction. These methods were developed and validated on brain MRI, on manually labeled cohorts with an agreed protocol~\citep{Klein2009}, which is the most mature benchmark setting for atlas construction. All of these construct the atlas with atlas-specific objectives.
In contrast, we obtain the atlas from a generator trained for synthesis alone, without any atlas objectives.

\paragraph{Learned templates in vision.}
Outside medicine, joint-alignment methods optimize an aligner together with a template of an image collection: congealing averages the aligned stack~\citep{learned2006data}, GANgealing aligns images to a GAN sample at a learned latent~\citep{peebles2022gan}, and neural congealing maps images into a joint atlas of semantic features~\citep{ofri2023neural}. For 3D shapes, deep implicit templates learn one implicit surface per category together with a per-shape deformation onto it, DIT through a learned warp field~\citep{zheng2021deep} and DIF-Net through a deformation-plus-correction field~\citep{deng2021deformed}. In every case, the template is a trained object, and its quality is measured under the aligner it was trained with. We train nothing for the template, and we evaluate it under a registration protocol it has never seen, alongside the templates these methods produce.

\subsection{Deterministic sampling and the role of noise}
Deterministic samplers such as DDIM and the probability-flow ODE \citep{song2020denoising,song2021scorebased} are designed to preserve the forward marginals, so each initialization still yields a distinct sample. Previous work~\citep{meng2021sdedit,huberman2024edit} edits the injected noise during the sampling steps to edit a sample, exploiting the fact that the noise carries individual variation. Our sampler is the opposite operation that discards the noise rather than steering it, and, unlike DDIM, it does not preserve the marginals. The identification of the denoiser with the posterior mean and the score is classical~\citep{efron2011tweedie}. To our knowledge, no prior work reports this exact convergence, or its use as an atlas.


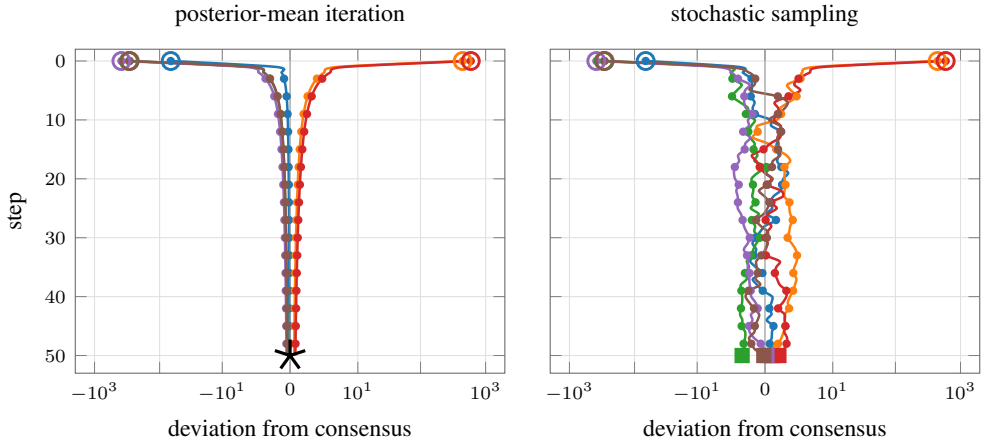
\begin{figure}[!b]
    \centering
    \input{figures/funnel/fig_funnel}
    \setlength{\abovecaptionskip}{0.5em}
    \setlength{\belowcaptionskip}{-1em}
    \caption{\textbf{Canonical convergence}. Six random initializations (circles) under the posterior-mean iteration contract onto one image (star) within the first steps; the same network sampled stochastically keeps them distinct (squares). Brain-T1 generator, first 50 of 900 steps.
    }
    \label{fig:deviation}
\end{figure}

\section{The Posterior-Mean Iteration and Canonical Convergence}
\label{sec:mechanism}

We use a trained denoising diffusion probabilistic model \citep{ho2020denoising}, with forward process $x_t=\sqrt{\bar\alpha_t}\,x_0+\sqrt{1-\bar\alpha_t}\,\epsilon$, $\epsilon\sim\mathcal N(0,I)$, and a network $\epsilon_\theta(x_t,t,y)$ trained to predict $\epsilon$ under an optional condition $y$ (\eg\ anatomy class or age). We do not modify training in any way; the only change is how the reverse process is run at inference.

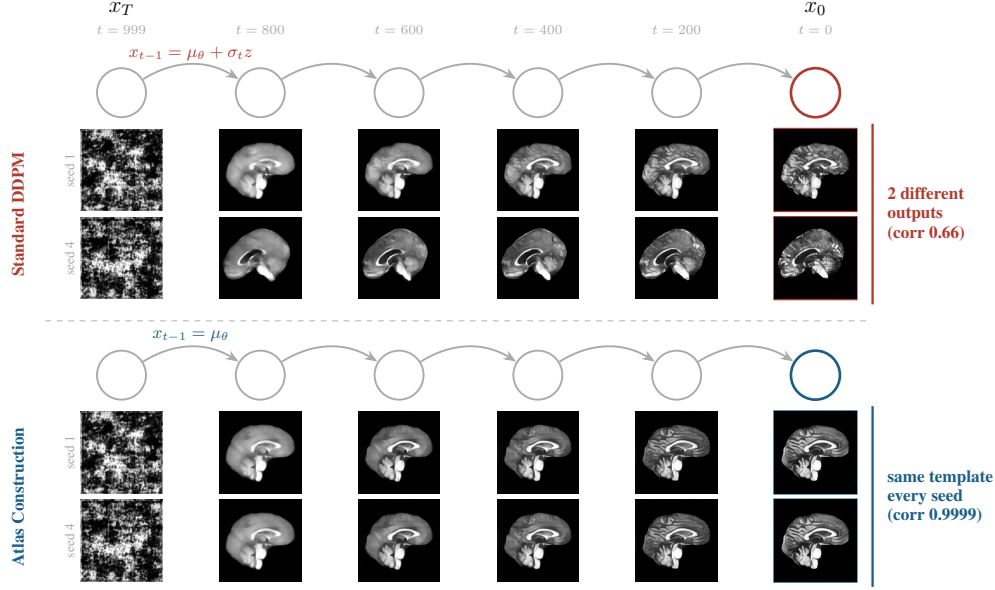
\begin{figure}[t]
\centering
\scalebox{.72}{\input{figures/fig_algo}}
\caption{
\textbf{Canonical convergence.} Reverse process of one pretrained brain-MRI diffusion model at six timesteps, decoding $\hat x_0=\mathbb E[x_0\mid x_t]$; each row uses an independent seed. \textbf{Top}: stochastic sampling, each seed a different brain (cross-seed correlation 0.66). \textbf{Bottom}: the posterior-mean iteration on the same network; every seed converges to the same image (0.9999).
}
\label{fig:main}
\end{figure}

\begin{wrapfigure}{r}{0.3\linewidth}
\vspace{-1em}
    \centering
    \includegraphics[width=\linewidth]{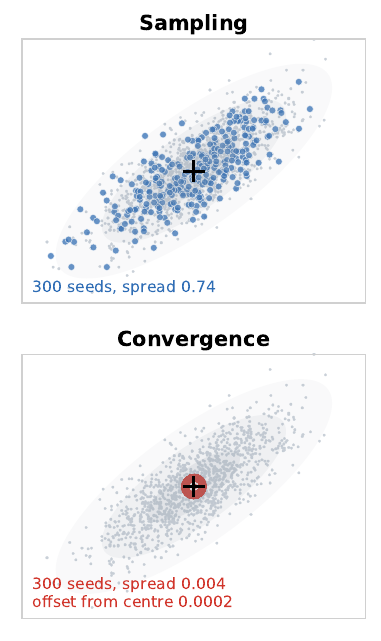}
    \setlength{\abovecaptionskip}{-1em}
    \setlength{\belowcaptionskip}{.5em}
    \caption{\textbf{Sampling versus convergence} on a population with a known center.}
    \label{fig:toy}
\end{wrapfigure}
\paragraph{The posterior-mean iteration.} Rather than sampling $p_\theta(x_{t-1}\mid x_t,y)$, we replace each stochastic transition by its conditional mean,
\begin{equation}
x_{t-1}\leftarrow\mu_\theta(x_t,t,y)=\mathbb E_{p_\theta(x_{t-1}\mid x_t,y)}[x_{t-1}],
\end{equation}
where ordinary sampling would add $\sigma_t z$, $z\sim\mathcal N(0,I)$, to this mean at every step.
This differs from DDIM, which re-inserts the predicted noise at the schedule's amplitude, $x_{t-1}=\sqrt{\bar\alpha_{t-1}}\,\hat x_0+\sqrt{1-\bar\alpha_{t-1}}\,\hat\epsilon$, and so preserves each initialization; the posterior-mean iteration re-inserts nothing, so the state is driven only by the denoiser's pull toward the population's center.


The mean update is determined by the denoised estimate $\hat x_0(x_t,t,y)$, itself linked to the score of the noised marginal via Tweedie's formula \citep{efron2011tweedie},
\begin{equation}
\label{eq:tweedie}
\hat x_0(x_t) = \tfrac{1}{\sqrt{\bar\alpha_t}}\big(x_t + (1-\bar\alpha_t)\nabla_{x_t}\log f_{X_t}(x_t)\big).
\end{equation}
$\hat x_0(x_t)=\mathbb E[x_0\mid x_t]$ is the expected clean image given the noisy state $x_t$. Without any injected noise, each update moves the state toward $\hat x_0$, the center of the population, from the current state.
\Cref{fig:toy} makes this concrete on a 2D Gaussian population, whose center and noised density $f_{X_t}$ are known. With a known score function $\nabla_{x_t}\log f_{X_t}(x_t)$ in place of a trained network, all $300$ seeds reach the center (offset $2{\times}10^{-4}$), while stochastic sampling from the same seeds spreads across the population. Convergence to the center is a property of the operator, not of a trained network.

\paragraph{Canonical convergence and the intrinsic atlas.} Let $\Phi_\theta(x_T;y)$ denote the endpoint obtained by running the posterior-mean iteration from the initialization $x_T$ under condition $y$; it is a deterministic function of $x_T$. We say the model exhibits \emph{canonical convergence} at $y$ if $\Phi_\theta(x_T;y)$ is the same for almost every $x_T\sim\mathcal N(0,I)$; that common endpoint, written $x^\star(y)$, is the model's \emph{intrinsic atlas} for condition $y$.
This behavior can be verified easily: if a population center exists, every random $x_T$ should return the same image; if not, the trajectories should scatter. As shown in~\Cref{fig:deviation}, comparing against the stochastic sampling, the posterior-mean iteration presents clear convergence.

\begin{figure}[t]
    \centering
    \captionsetup[subfigure]{skip=1pt} 
    \subfloat[3D Chest, CT]{
        \includegraphics[width=0.32\textwidth]{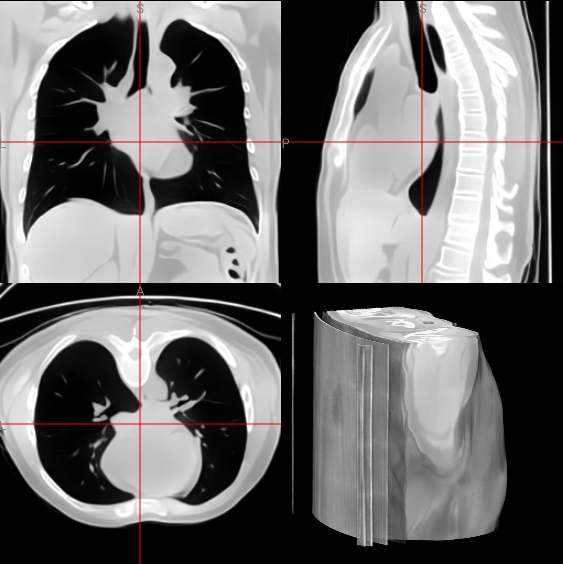}
    }
    \subfloat[3D Legs, CT]{
        \includegraphics[width=0.32\textwidth]{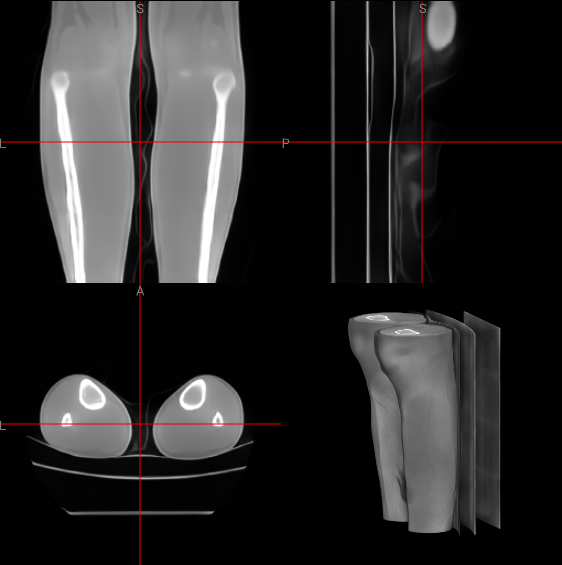}
    }
    \begin{minipage}[b]{0.292\textwidth}
        \subfloat[2D Fundus]{
            \includegraphics[width=\textwidth,trim={20.7cm 24.2cm 0cm 0},clip]{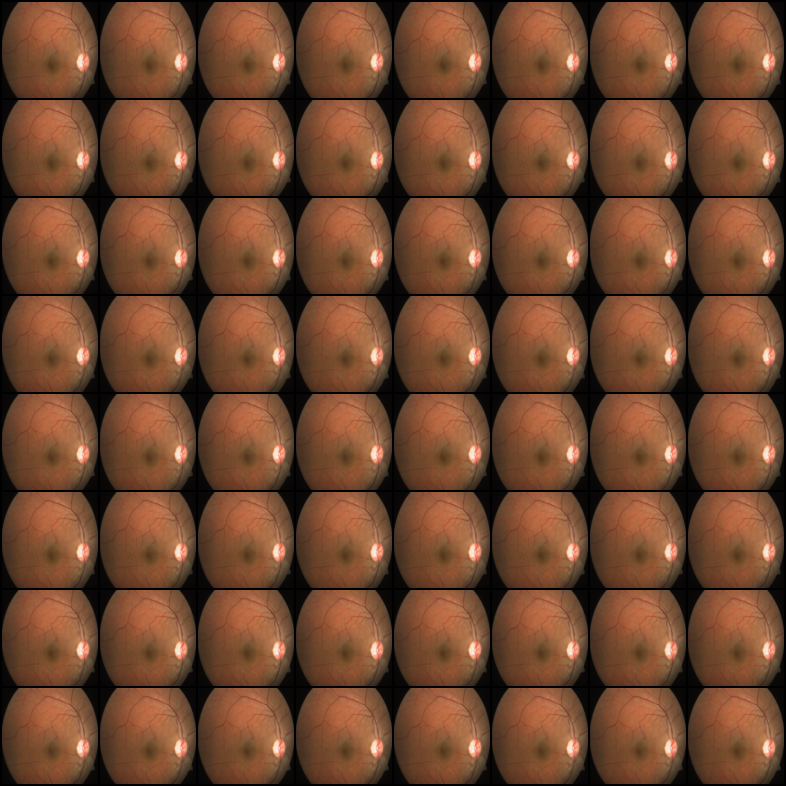}
        }
        \\[-0.1em]
        \subfloat[2D Chest XRay]{
            \includegraphics[width=\textwidth,trim={27.7cm 32.2cm 0cm 0},clip]{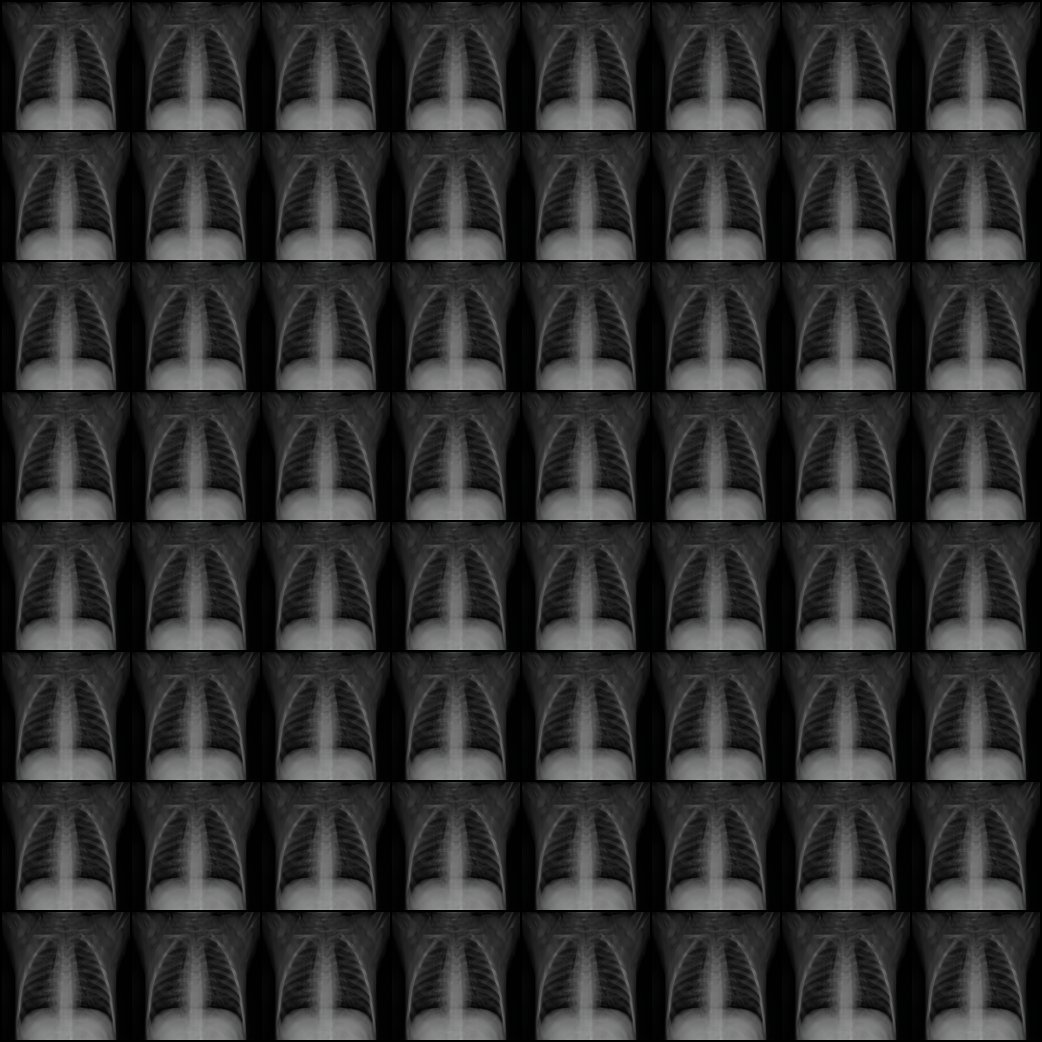}
        }
    \end{minipage}
    
    \subfloat[3D Airplane, ModelNet]{
        \includegraphics[width=0.32\textwidth,trim={1.5cm 9.7cm 29.5cm 3cm},clip]{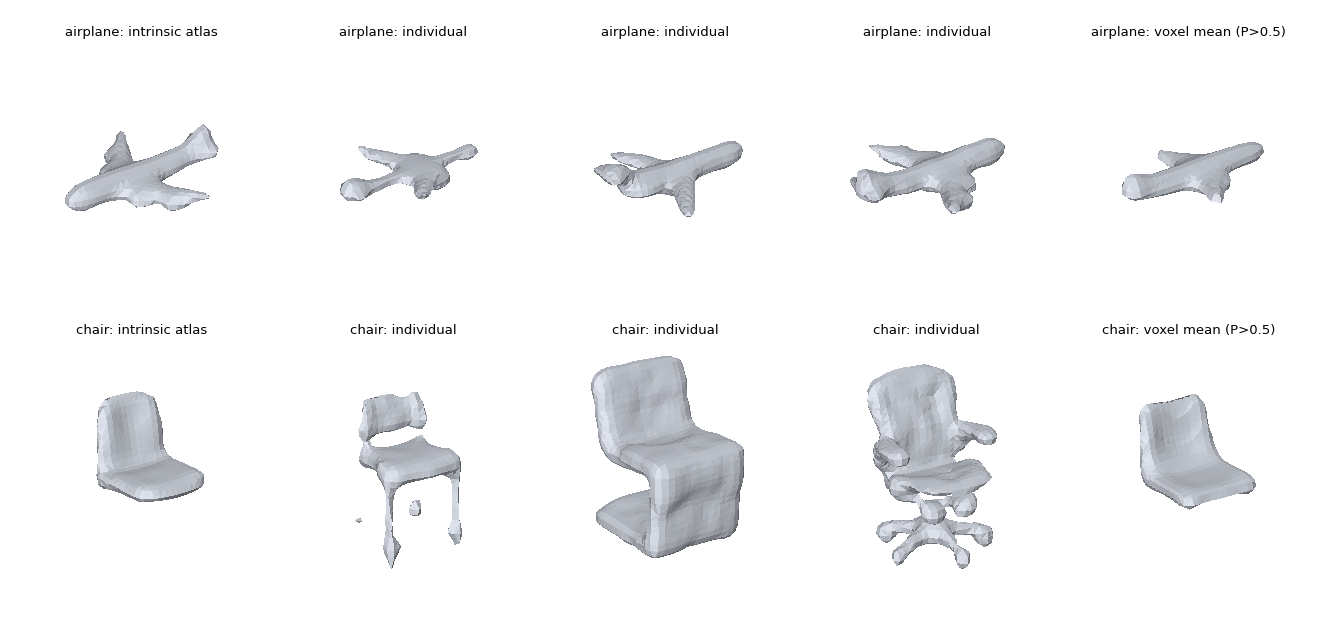}
    }
    \subfloat[3D Chair, ModelNet]{
        \includegraphics[width=0.32\textwidth,trim={1.5cm 2.3cm 29.5cm 10cm},clip]{figures/shapes_view2.png}
    }
    \begin{minipage}[b]{0.292\textwidth}
        \subfloat[2D Face, FFHQ]{
            \includegraphics[width=.495\textwidth,trim={.7cm 11cm 28cm 1.5cm},clip]{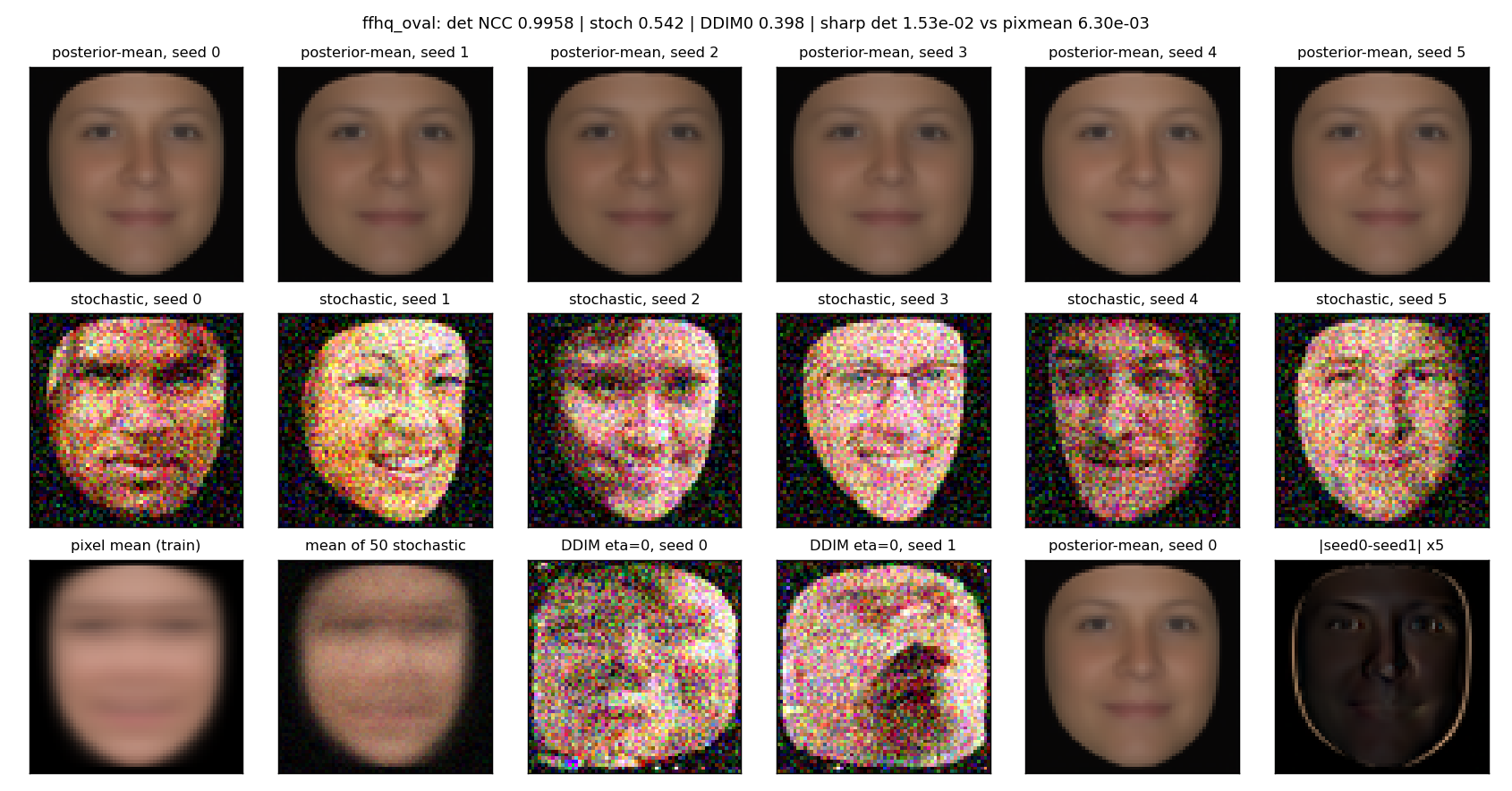}
            \includegraphics[width=.495\textwidth,trim={.7cm 11cm 28cm 1.5cm},clip]{figures/ffhq_oval_recovered.png}
        }
        \\[-0.1em]
        \subfloat[2D Letter ``A"]{
            \includegraphics[width=.495\textwidth,trim={.7cm 11cm 28cm 1.5cm},clip]{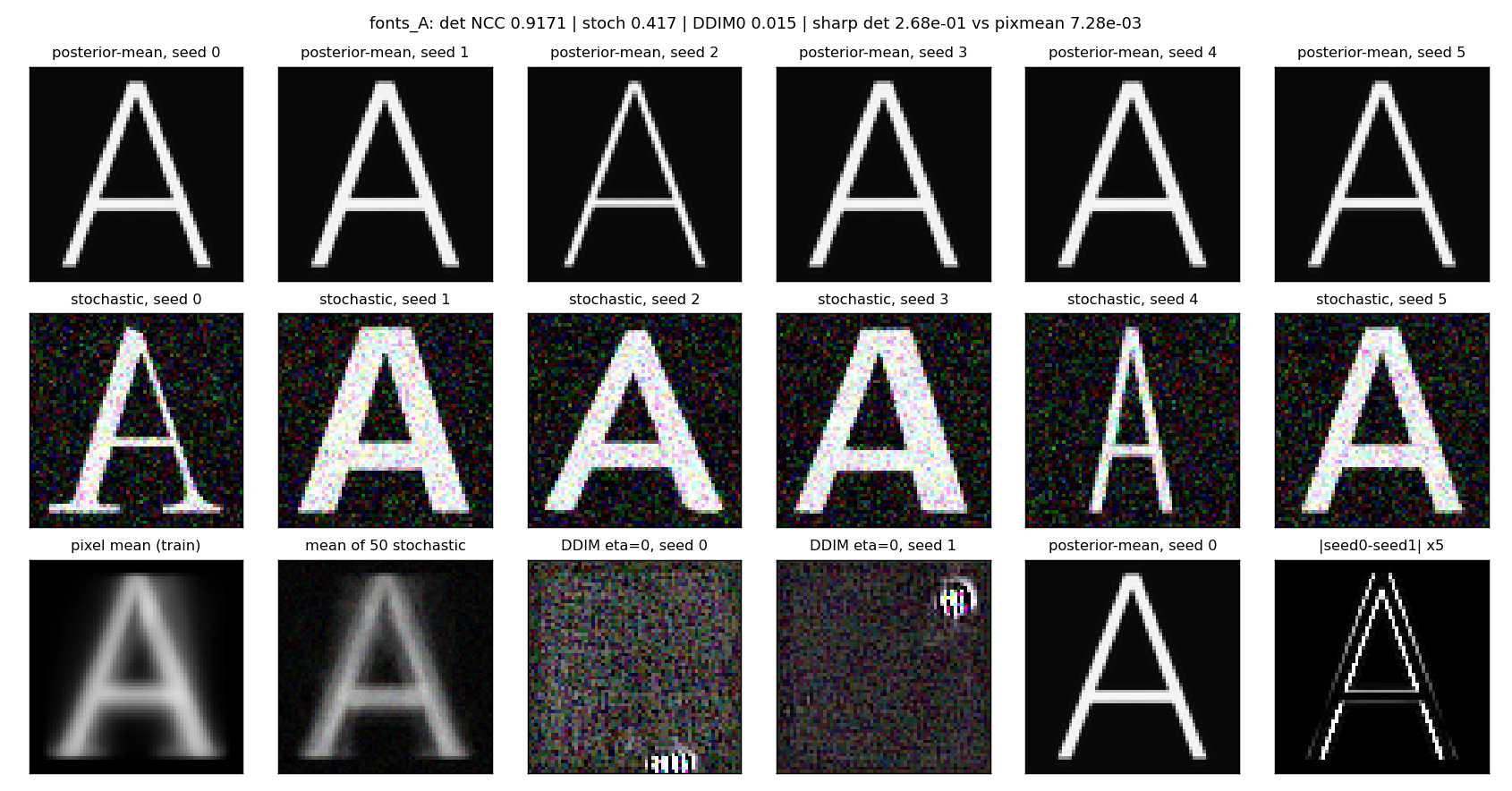}
            \includegraphics[width=.495\textwidth,trim={.7cm 11cm 28cm 1.5cm},clip]{figures/fonts_A_recovered.png}
        }
    \end{minipage}
    \setlength{\belowcaptionskip}{-1em}
    \caption{\textbf{Generality.} Intrinsic atlases on medical (top) and non-medical (bottom) populations, in 3D and 2D: (a) chest CT, (b) leg CT, (c) retinal fundus, (d) chest X-ray, (e) ModelNet airplane, (f) ModelNet chair, (g) FFHQ faces, (h) the letter ``A'' across typefaces.}
    \label{fig:generality}
\end{figure}

\paragraph{Atlas Visual Quality Across Modalities and Domains.}
\label{sec:generality}
As shown in \cref{fig:generality}, we present intrinsic atlases from multiple domains, including chest and leg CT, retinal fundus, CelebA faces, the letter ``A'' across $3{,}796$ typefaces, and ModelNet airplanes and chairs, each with cross-seed SSIM $\geq0.99$. The atlas keeps what the population agrees on and drops what it does not: the ``A'' is a clean sans-serif glyph from a mix of sans, serif, and script faces. Interestingly, the chair atlas has a seat and backrest but no legs, since the population does not share a leg design.

\section{The Utility of the Intrinsic Atlas}
\label{sec:results}
Atlas construction has been developed and benchmarked almost entirely on brain MRI: the classical methods~\citep{joshi2004unbiased,AVANTS200826,fonov2011unbiased}, the reference evaluation of registration algorithms~\citep{Klein2009}, and every learned-template method we compare against~\citep{Dalca2019learning,dey2021generative,ding2022aladdin,abulnaga2025multimorph} were designed and validated there. We therefore adopt this well-benchmarked domain as our primary evaluation, on IBSR18~\citep{ibsr}, Mindboggle101~\citep{mindboggle101}, and ABIDE-I~\citep{di2014autism}, following its established protocols~\citep{Klein2009,Dalca2019learning,abulnaga2025multimorph,hering2022learn2reg}.
Meanwhile, we evaluate on domains for template correspondence, namely CelebA~\citep{liu2015faceattributes} faces, Montgomery~\citep{jaeger2014two} chest X-rays, and 3D KeypointNet~\citep{you2020keypointnet} chairs and cars. Since those domains have no anatomical segmentations, we report the percentage of correct keypoints (PCK) and the mean/median label transfer IoU, following~\citep{peebles2022gan,deng2021deformed}. Details are provided in the supplementary material.

\paragraph{Registration Utility of the Intrinsic Atlas.} The intrinsic atlas is best or second in every dataset on each domain. On brain MRI, MultiMorph has a higher Dice by at most half a point on Mindboggle101 and ABIDE ($+0.004$ each), and by three points on IBSR18, while ours is the most central and most regular template on every cohort. The same best or second trend persists in other domains. VoxelMorph is first on faces ($0.694$ PCK@0.1) but fourth on chairs ($0.335$, $0.13$ behind). Aladdin is first on chest X-ray ($0.727$) but last on faces ($0.583$, below the pixel mean). Our method is stronger than DIF-Net and DIT in PCK on both 3D shapes, and best on both PCK@0.05 and part-label IoU on 3D cars, second to DIF-Net by 0.001 on PCK@0.1.

\newcommand{\best}[1]{\textbf{{#1}}}
\newcommand{\second}[1]{{\underline{#1}}}
\begin{table}[t]
\centering\footnotesize
\setlength{\abovecaptionskip}{.25em}
\caption{\textbf{Atlas construction evaluation across domains.} Registration is ANTs in every domain, with one preset per domain (SyNQuick for brain MRI, SyNCC for faces and X-ray, SyN for shapes). Within a domain, the same registration protocol is used for every template, so only the template varies. Every cell is the mean of at least 3 repeated runs.}
\label{tab:pck_ants}

\begin{subtable}[b]{\textwidth}
\setlength{\tabcolsep}{1.2pt}
\centering
\setlength{\abovecaptionskip}{0em}
\setlength{\belowcaptionskip}{.25em}
\caption{\textbf{On T1 Brain MRI.} \textbf{Metrics}: We report Dice, the accuracy criterion of the reference benchmark~\citep{Klein2009}, plus three deformation metrics such as centrality~\citep{Dalca2019learning,abulnaga2025multimorph}, mean per-subject displacement $|u|$~\citep{Dalca2019learning}, and SDlogJ~\citep{hering2022learn2reg}. \textbf{Models}: For T1 MRI, we report two intrinsic atlases, one from the off-the-shelf model~\citep{khader2022medical} and one from our age-conditioned model of \cref{sec:age}, fine-tuned with the age condition disabled. Neither model is trained on  IBSR18, Mindboggle101, or ABIDE.}
\begin{tabular}{l cccc cccc cccc}
\toprule
& \multicolumn{4}{c}{IBSR18} & \multicolumn{4}{c}{Mindboggle101} & \multicolumn{4}{c}{ABIDE} \\
\cmidrule(lr){2-5}\cmidrule(lr){6-9}\cmidrule(lr){10-13}
 & Dice$\uparrow$ & Cent.$\downarrow$ & $|u|\downarrow$ & SDlogJ$\downarrow$
 & Dice$\uparrow$ & Cent.$\downarrow$ & $|u|\downarrow$ & SDlogJ$\downarrow$
 & Dice$\uparrow$ & Cent.$\downarrow$ & $|u|\downarrow$ & SDlogJ$\downarrow$ \\
\midrule
\multicolumn{7}{l}{\color{darkgray}\textit{\textbf{Baseline Methods}}} \\
\quad voxel mean            
    & 0.723 & 2.65 & 3.40 & 0.084 
    & 0.507 & 2.85 & 3.65 & 0.086 
    & 0.665 & 2.59 & 3.21 & 0.076 \\
\quad ANTs 
    & 0.722 & 2.80 & 3.47 & 0.070 
    & 0.521 & 3.03 & 3.78 & 0.080 
    & 0.669 & 2.75 & 3.34 & \second{0.064} \\
\midrule
\multicolumn{7}{l}{\color{darkgray}\textit{\textbf{Deep learning methods}}} \\
\quad VoxelMorph             
    & 0.732 & 2.24 & 3.07 & 0.071 
    & 0.527 & 2.59 & 3.49 & 0.086 
    & 0.674 & 2.16 & 2.92 & 0.065 \\
\quad Aladdin                
    & 0.730 & 2.82 & 3.56 & 0.082 
    & 0.512 & 3.23 & 4.01 & 0.093 
    & 0.664 & 2.67 & 3.32 & 0.075\\
\quad MultiMorph $*$         
    & \best{0.777} & 1.70 & 2.83 & 0.077 
    & \best{0.546} & 2.40 & 3.47 & 0.094 
    & \best{0.691} & 1.67 & 2.77 & 0.072 \\
\midrule
\quad \textbf{Ours} (off-the-shelf)
    & 0.733 & \second{1.65} & \second{2.59} & \second{0.062}
    & 0.529 & \second{1.59} & \second{2.74} & \best{0.065}
    & {0.675} & \second{1.58} & \best{2.43} & \best{0.055} \\
\quad \textbf{Ours}
    & \second{0.742} & \best{1.52} & \best{2.53} & \best{0.060}
    & \second{0.542} & \best{1.41} & \best{2.72} & \second{0.066}
    & \second{0.687} & \best{1.53} & \best{2.43} & \second{0.056} \\
\bottomrule
\multicolumn{7}{l}{\footnotesize $*$: established atlases, without training on the same dataset.} \\
\end{tabular}
\end{subtable}

\begin{subtable}[b]{\textwidth}
\centering
\small
\setlength{\tabcolsep}{1.5pt}
\setlength{\abovecaptionskip}{0em}
\setlength{\belowcaptionskip}{.25em}
\caption{\textbf{On other domains}. \textbf{Metrics}: PCK@0.1\,/\,@0.05$\uparrow$ is the fraction of landmarks transferred through the template that land within $0.1$\,/\,$0.05$ of the image size. IoU avg/mid denotes the mean\,/\,median of the part segmentation (e.g.\ seat, back, leg) by label transfer~\citep{deng2021deformed}; 5 training data are deformed to its template, then deforming another data to validate the matches by nearest-neighbor voting. \textbf{Models}: generators trained by us on each domain's training split.}
\begin{tabular}{lcc cc cc}
\toprule
 & CelebA (faces) & Mont. (x-ray) & \multicolumn{2}{c}{KeypointNet (Chair)} & \multicolumn{2}{c}{KeypointNet (Car)} \\
\cmidrule(lr){2-2}\cmidrule(lr){3-3}\cmidrule(lr){4-5}\cmidrule(lr){6-7}
&  PCK{\scriptsize @0.1/@0.05}
& PCK{\scriptsize @0.1/@0.05}
& PCK{\scriptsize @0.1/@0.05} & IoU {\scriptsize avg/mid}
& PCK{\scriptsize @0.1/@0.05} & IoU {\scriptsize avg/mid}\\
\midrule
\multicolumn{3}{l}{\color{darkgray} \textit{\textbf{Baseline Methods}}} \\
\quad pixel/voxel mean 
    & 0.613\,/\,0.396
    & 0.675\,/\,0.455
    & 0.252\,/\,0.051  & 0.603\,/\,0.605
    & 0.739\,/\,0.333  & 0.615\,/\,0.626 \\
\quad ANTs   
    & 0.655\,/\,0.472 
    & 0.660\,/\,0.429
    & -- & --
    & -- & --\\
\midrule
\multicolumn{3}{l}{\color{darkgray} \textit{\textbf{Deep Learning methods}}} \\
\quad VoxelMorph & \best{0.694}\,/\,\best{0.489}
            & {0.644}\,/\,{0.424} 
            & 0.335\,/\,0.079  & 0.644\,/\,0.638
            & 0.722\,/\,0.311  & 0.614\,/\,0.624 \\
\quad Aladdin    & 0.583\,/\,0.362 
            & \best{0.727}\,/\,\best{0.514} 
            & 0.309\,/\,0.073  & 0.605\,/\,0.616 
            & 0.731\,/\,0.329  & {0.619}\,/\,\second{0.635} \\
\quad DIF-Net    & -- & -- 
            & \second{0.448}\,/\,\second{0.114}  & 0.668\,/\,0.656 
            & \best{0.755}\,/\,\second{0.351}  & \second{0.621}\,/\,0.632 \\
\quad DIT        & -- & -- 
            & 0.270\,/\,0.063  & 0.595\,/\,0.585 
            & 0.736\,/\,0.325  &0.618\,/\,0.631 \\
\midrule
\quad \textbf{Ours} & \second{0.658}\,/\,\second{0.438} 
        & \second{0.685}\,/\,\second{0.429} 
        & \best{0.465}\,/\,\best{0.117}  & \best{0.696}\,/\,\best{0.679} 
        & \second{0.754}\,/\,\best{0.355}  & \best{0.626}\,/\,\best{0.639} \\
\bottomrule
\end{tabular}
\end{subtable}

\end{table}

\section{On-Demand Atlas Families via Conditioning}
\label{sec:age}

If the intrinsic atlas is the central anatomy of the population, the model learned under condition $y$, then varying $y$ should yield a whole family of intrinsic atlases from a single trained model.
We test this with a continuous, clinically central conditioning variable: age. We fine-tune a brain-T1 generator with a continuous age condition on $899$ T1 MRIs (IXI, OASIS-1; cognitively normal subjects, ages $18$--$94$), and recover the atlas at a fixed age value with the same deterministic procedure, at guidance scale $1$, with no per-age optimization. See supplementary for technical details.

\begin{figure}[t]
\centering
\includegraphics[width=.95\linewidth]{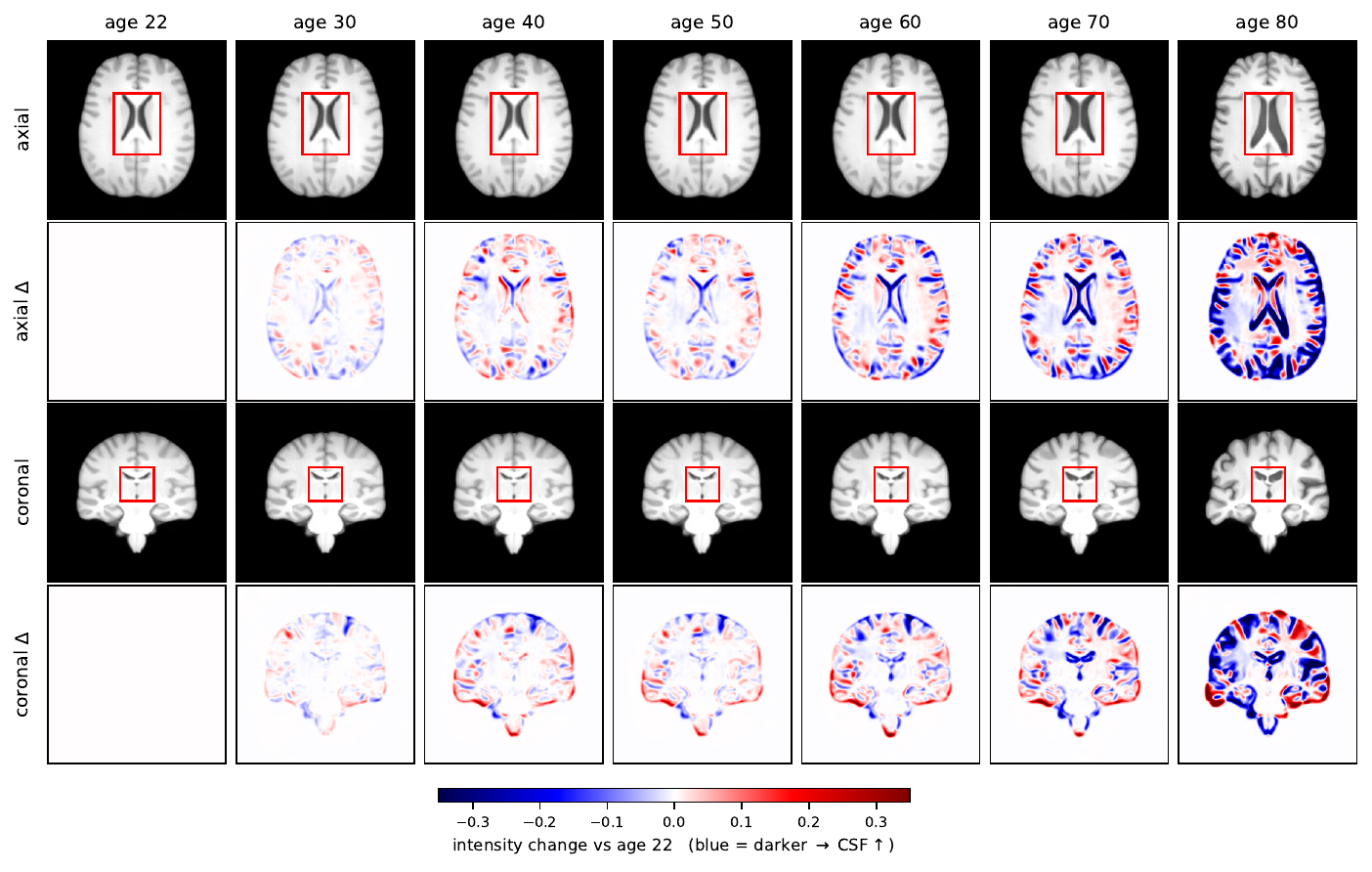}
\setlength{\abovecaptionskip}{0em}
\setlength{\belowcaptionskip}{-1em}
\caption{\textbf{The age-conditioned family, visually.} Columns are references recovered for ages $22$--$80$ by varying only the age condition. Grayscale rows are the intrinsic atlases; rows beneath show the voxelwise difference from the youngest atlas on a shared diverging scale. Boxes (fixed across ages) frame the lateral ventricles; their enlargement is visible directly in grayscale.}
\label{fig:age_sweep_app}
\end{figure}

\begin{figure}[h]
\centering
\begin{subfigure}[t]{0.49\textwidth}
\includegraphics[width=\linewidth]{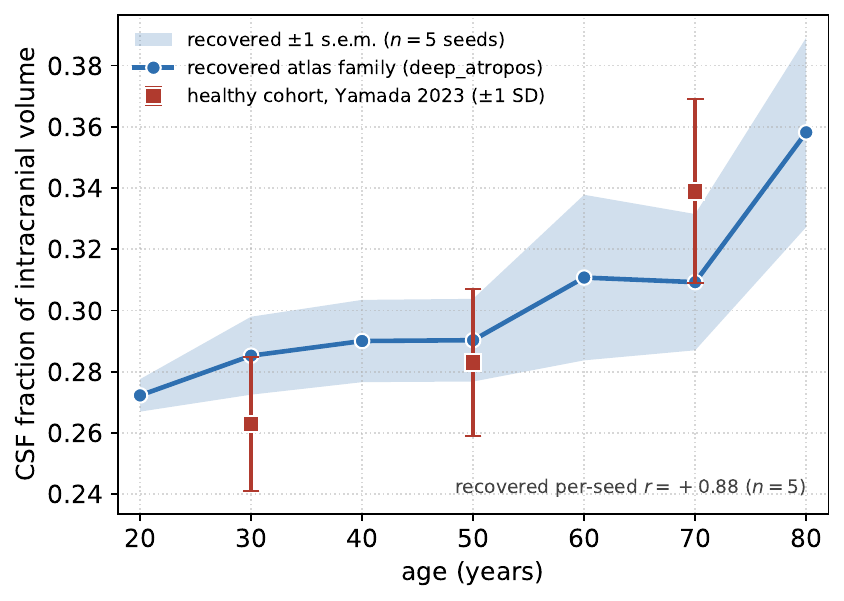}
\setlength{\abovecaptionskip}{-.5em}
\setlength{\belowcaptionskip}{-.5em}
\caption{\textbf{The age-conditioned family reproduces the CSF-expansion signature of healthy aging.} Blue: CSF fraction vs.\ age (mean $\pm$ s.d., $n{=}5$ seeds). Red: independent reference values from \citet{yamada2023aging}, aligned by one disclosed constant shift.}
\label{fig:age_aging}
\end{subfigure}
\hfill
\begin{subfigure}[t]{0.49\textwidth}
\includegraphics[width=\linewidth]{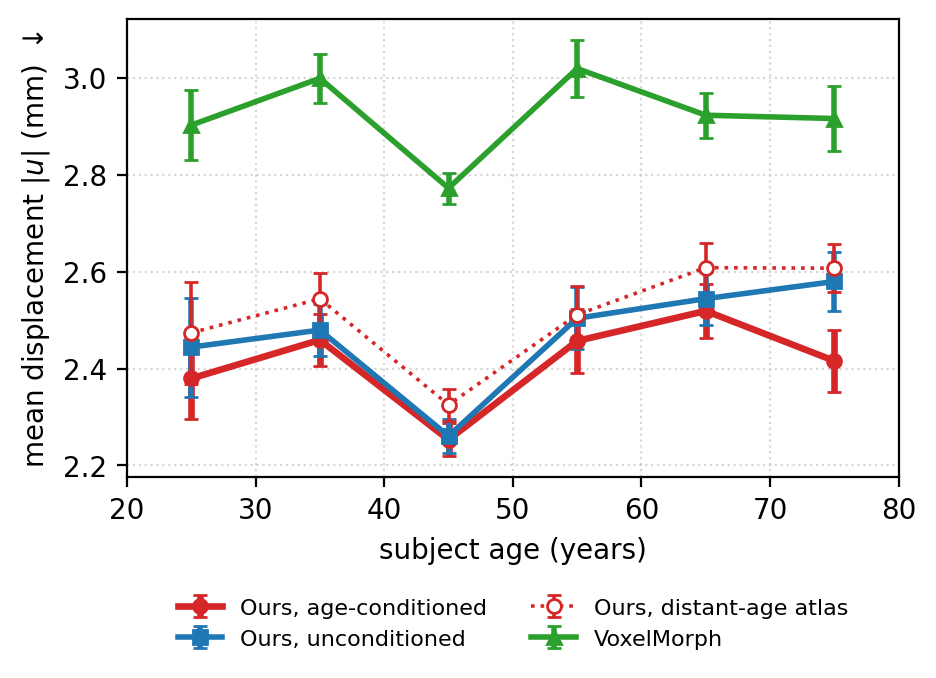}
\setlength{\abovecaptionskip}{-.5em}
\setlength{\belowcaptionskip}{-.5em}
\caption{\textbf{Age-conditioned vs. unconditioned atlases}. Mean displacement $|u|$ on $78$ subjects, grouped by decade. Age-matched atlases show a consistent gain.
}
\label{fig:age_utility}
\end{subfigure}
\end{figure}

\paragraph{Biological validity.}
An age-conditioned atlas family is only meaningful if it reproduces known age effects. One of the well-established effects in healthy aging is brain atrophy: the ventricles enlarge, and the cerebrospinal-fluid (CSF) fraction of intracranial volume rises monotonically and increasingly steeply with age~\citep{resnick2003longitudinal,fjell2010structural}.
We segment each recovered atlas with a contrast-robust deep-learning segmenter and track the CSF volume fraction, the canonical marker of brain aging.
\Cref{fig:age_sweep_app} shows clear progressive ventricular enlargement during aging\footnote{We show ages $22$--$80$ to exclude the sparsely populated extremes of the $18$--$94$ training range.}, as annotated in red squares in the figure.
The recovered family reproduces the accelerating, monotonic CSF-expansion trend reported for cognitively normal cohorts \citep{yamada2023aging}, consistently across five independent noise seeds (per-seed correlation $r{=}{+}0.88$, spread ${<}0.2$ percentage points at every age).
Note that \citet{yamada2023aging} report values from a commercial volumetry pipeline, whereas ours come from an open-source segmenter~\citep{Tustison2021} with its own tissue boundaries, which produces a constant offset between the two curves. We therefore, in~\Cref{fig:age_aging}, compare the trend after removing a single constant shift.

\paragraph{Registration utility.} We test whether a conditioned atlas is a better registration target, as shown in~\cref{fig:age_utility}, on the $78$ IXI subjects of the cohort, aged $20$--$79$ (the $45$ OASIS-1 subjects are excluded). Compared to the non-age-conditioned atlas (blue line), the age-matched atlas (solid red line) requires less deformation for $53/78$ subjects and has the lower mean in every decade  ($-0.05$\,mm, $p{=}4{\times}10^{-5}$, $p<0.001$). The VoxelMorph template trained on the same data lies a further $0.4$--$0.5$\,mm above it ($78/78$, $p{=}2{\times}10^{-14}$).

As a control, each subject is also registered to an atlas at a distant age (dotted red line, $25$ for subjects aged $50$ or older, $75$ for younger ones). If conditioning carried no information, both would register equally well. Instead, the age-matched atlas gives a lower displacement error for $57$ of $78$ subjects, statistically significant ($p{=}5{\times}10^{-7}$, $p<0.001$), and its mean displacement lies below the distant-age atlas in every decade. Conditioning on age produces atlases specific to that age, not merely plausible variants.

\section{What Convergence Reveals About the Population}
\label{sec:failures}

\begin{figure}[!b]
    \tiny
    \centering
    \setlength{\tabcolsep}{1pt}
    \begin{tabular}{cc}
            \begin{subfigure}{0.46\textwidth}
                \centering
                \setlength{\tabcolsep}{1pt}
                \begin{tabular}{ccc}
                    {\footnotesize \makecell{Mode 1}}
                    & {\footnotesize \makecell{Mode 2}}
                    & {\footnotesize \makecell{Mode 3}}
                    \\
                    \includegraphics[width=.33\linewidth]{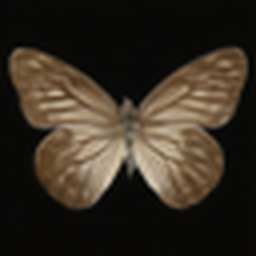}
                    & 
                    \includegraphics[width=.33\linewidth,trim={0 0 0 0.01cm},clip]{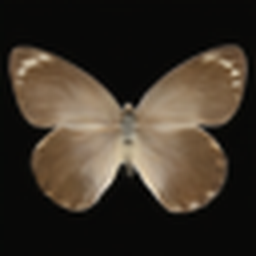}
                    & 
                    \includegraphics[width=.33\linewidth,trim={0 0 0 0.05cm},clip]{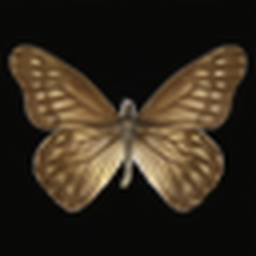}
                \end{tabular}
                \caption{
                    Multiple population consensus.
                }
                \label{fig:limitation_modalities_a}
            \end{subfigure}
            \hspace{.5em}
            \begin{subfigure}{0.46\textwidth}
                \centering
                \setlength{\tabcolsep}{1pt}
                \begin{tabular}{ccc}
                    {\footnotesize \makecell{$32\times 32$}}
                    & {\footnotesize \makecell{$64\times 64$}}
                    & {\footnotesize \makecell{$128\times 128$}}
                    \\
                    \includegraphics[width=.33\linewidth,trim={8.4cm 8.4cm 0cm 0.01cm},clip]{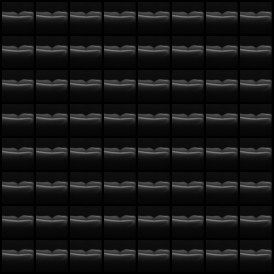}& 
                    \includegraphics[width=.33\linewidth,trim={16.4cm 16.4cm 0.1cm 0.1cm},clip]{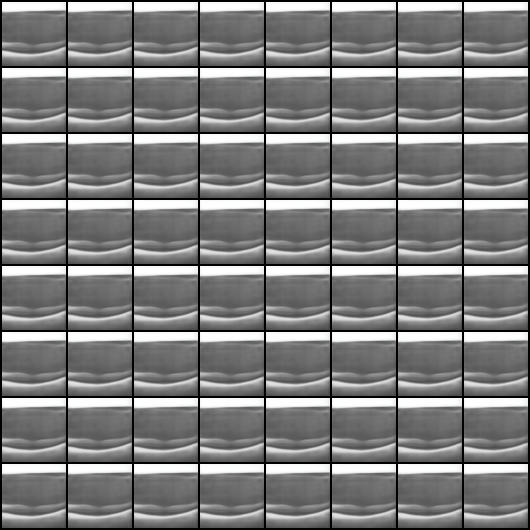}& 
                    \includegraphics[width=.33\linewidth,trim={24.2cm 24.2cm 0cm 0},clip]{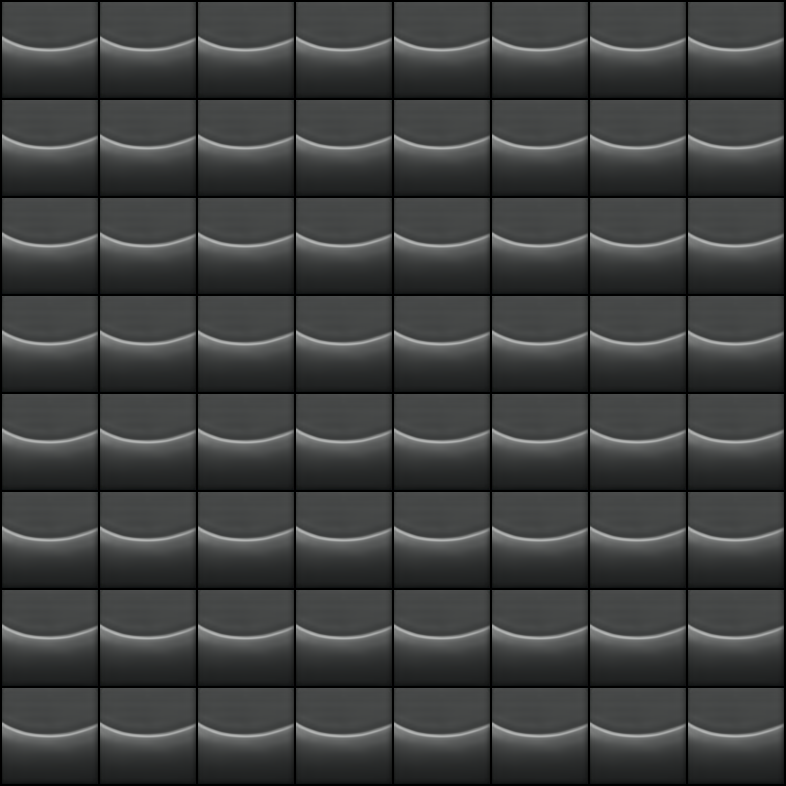}
                \end{tabular}
                \caption{Scale-dependent population consensus}
                \label{fig:limitation_modalities_b}
            \end{subfigure}
    \end{tabular}
    
    \begin{subfigure}{0.98\textwidth}
        \centering
        \begin{tikzpicture}
            \node [
                inner sep=5pt, 
                rounded corners=5pt
              ] at (-6.7,-.1) {
                \includegraphics[width=.78\linewidth,trim={0 0 0 0},clip]{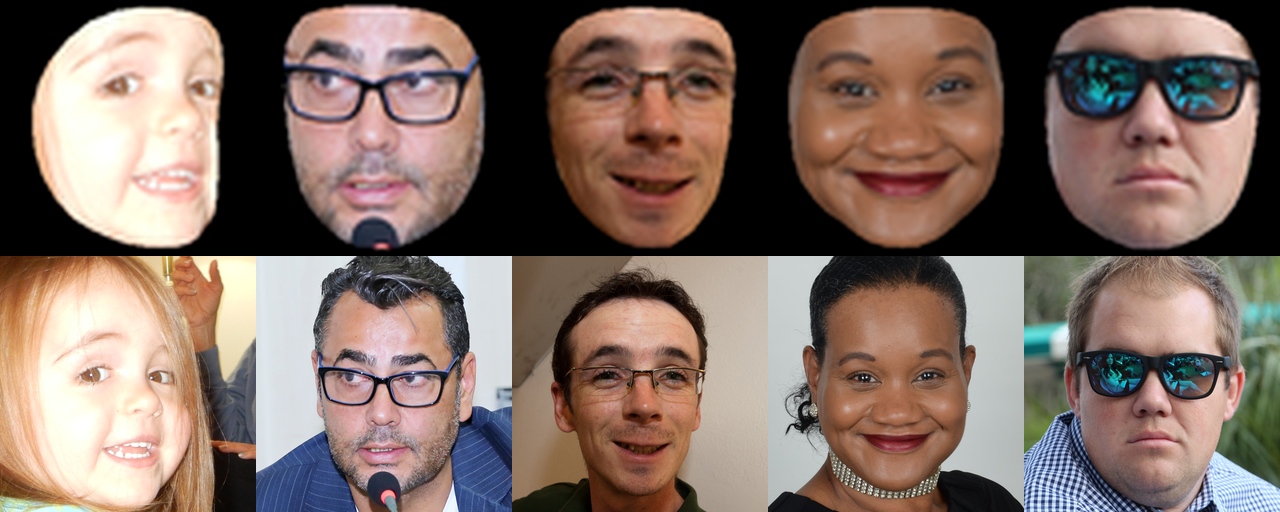} 
              };
          \node [
            inner sep=3pt, 
            shading=axis,
            align=center,
            left color=blue!15,
            right color=blue!15,
            rounded corners=3pt
          ] {
            \scriptsize Intrinsic Atlas \\
            \includegraphics[width=0.158\linewidth,trim={0 0 0 0.22cm},clip]{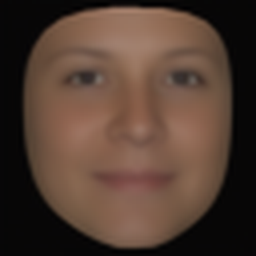} \\
            \includegraphics[width=0.158\linewidth,trim={0 0 0 0.13cm},clip]{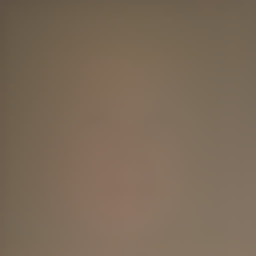}
          };
        \end{tikzpicture}
        \setlength{\abovecaptionskip}{0em}
        \caption{\textbf{A remedy to restore the population consensus.} The same five FFHQ subjects, cropped to the face oval (top) and original (bottom), with the intrinsic atlas each population yields (right): a consensus structure gives a clear population atlas; a blurred image otherwise.}
        \label{fig:limitation_modalities_c}
    \end{subfigure}

    \caption{
    \textbf{What canonical convergence reveals.} \textbf{(a)} Multiple consensus: Pieridae seeds resolve to one of three pattern variants. \textbf{(b)} Scale-dependent consensus: OCT is coherent at $32{\times}32$ and degrades with resolution. \textbf{(c)} A remedy: the original frames (bottom) share no consensus and yield a blur; cropping the same subjects to the face oval (top) restores the consensus and a sharp atlas.
    }
    \label{fig:failures_overview}
\end{figure}

Our method always converges on the population's consensus, in whatever state that consensus exists. \Cref{fig:failures_overview} shows the three forms we observe when it is not a single sharp template.

\textbf{Multiple templates.} When a population contains distinct sub-groups, the initialization selects among them. Within one butterfly family of Pieridae, $60$ seeds resolve to one of three similar pattern variants. \cref{fig:limitation_modalities_a} shows three generated intrinsic atlases of the patterned sub-groups.

\textbf{Scale-dependent template.} A population can share structure at one modeling scale but not another: retinal OCT data contains a very simple structure, and its intrinsic atlases are coherent at each resolution but vary across resolutions, as shown in~\cref{fig:limitation_modalities_b}. Though every intrinsic atlas makes sense pixel-wise, the lower-resolution $32\times 32$ atlas presents a more medically meaningful retinal structure with a clear fovea. Coherence is thus a pixel-space consensus, and the modeling scale decides which structures take part in it.

\textbf{No template.} When members vary in much of their content, the population consensus may not be a meaningful structure. As shown in~\cref{fig:limitation_modalities_c}, with a facial dataset, FFHQ, trained on original frames, where hair, background, and clothing vary, the convergence results in a featureless image. However, by manually removing those noises, the population-level consensus can be clearly reached.

\section{Discussion}
\label{sec:discussion}


Classical construction computes a Fr\'echet mean, a statistical center defined by minimizing summed deformation under a chosen deformation model. The intrinsic atlas instead follows the generator's density, with no deformation model. By Tweedie's formula in~\cref{eq:tweedie}, each step of the ascent equals the mean displacement, $\mathbb E[x_0 - x \mid x]$, from the current expectation $x$ to the population. The population itself, through the generator's density, defines the center. This is why the intrinsic atlas is central by construction, and the most central under registration in~\Cref{tab:pck_ants}. 
\vspace{-.8em}
\begin{figure}[h]
    \centering
    \includegraphics[width=\linewidth]{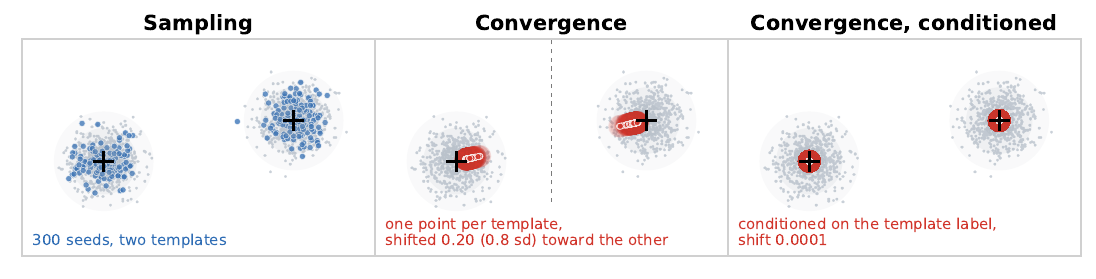}
    \setlength{\abovecaptionskip}{-1.5em}
    \setlength{\belowcaptionskip}{-1em}
    \caption{\textbf{Sampling versus convergence} on a population with two centers. \textit{Left}: stochastic samples cover both. \textit{Middle}: non-conditioned convergence finds two centers, but they pull on each other. \textit{Right}: conditioned convergence finds two separated centers.}
    \label{fig:toy_two}
\end{figure}

\paragraph{A structural bias, with a structural fix.}
If a population contains multiple centers, as shown in~\cref{fig:toy_two}, each recovered center is drawn toward the others, and conditioning
restores their centering. The age family shows the same effect. The unconditioned atlas resembles the age-$45$ atlas most, close to the mean training age of $47$, so it is already the atlas of the subjects in their $40$s, and the corresponding conditioning shows a minimal gain. As shown in~\cref{fig:age_utility}, the further a subject's age is from that mean, the more the conditioned atlas gains, least in the $40$s and most in the $70$s.

As the three Pieridae variants of~\cref{fig:limitation_modalities_a} show, seeds splitting into distinct images is itself a diagnostic that a population has several centers, with no label needed. The same principle behind mean-shift's use of multiple starts to find a density's modes~\citep{comaniciu2002mean}.
A mode-detection tool, with a principled agreement threshold, is left for future work.


\paragraph{From a registration target to registration.}
The intrinsic atlas is a valid registration target, but the deformation onto it is still computed outside the generator. Prior congealing pipelines~\citep{learned2006data,peebles2022gan,ofri2023neural} require training to obtain correspondence.
Diffusion-based works~\citet{tang2023emergent,luo2023diffusion,kim2022diffusemorph} explored feature correspondence within diffusion models.
We show that the template needs no training. If the correspondence does not either, congealing reduces entirely to inference on a generator built only to synthesize.


\section{Conclusion}
\label{sec:conclusion}

We set out from a hypothesis about the reverse process of a diffusion model: it combines a denoiser pulling toward the population and injected noise selecting an individual, so removing the noise should isolate what the population shares, its template in the sense of pattern theory. 
The recovered template, the \textit{intrinsic atlas}, is an atlas in the practical sense: best or second-best on every dataset across brain MRI, faces, chest X-ray, and 3D shapes. Conditioning turns it into
a family on demand, matched to any covariate the generator carries, and where a population holds several templates or none, the same process resolves into one image per template, or into a blurry image if none exists. Atlas construction, conventionally a dedicated optimization repeated for every population and covariate, is reframed as a byproduct of generative modeling: wherever a population coheres, and a generator has been trained on it, its atlas is already inside.



\label{app:mainend}
\bibliography{references}
\bibliographystyle{iclr2026_conference}

\newpage
\appendix
\input{appendix}

\end{document}

%% file: figures/fig_algo_preamble.tex
\usetikzlibrary{arrows.meta,positioning,calc}
\graphicspath{{figures/markov_thumbs/}}
\definecolor{ddpmcol}{HTML}{B03A2E}
\definecolor{ourscol}{HTML}{1A5E8A}
\definecolor{lgry}{HTML}{AAAAAA}
\def\dx{2.55}
\def\tw{15mm}
\def\rowsep{1.62}
\def\thdrop{1.42}
\newcommand{\ts}[1]{\ifcase#1 999\or 800\or 600\or 400\or 200\or 0\fi}
\newcommand{\chainblock}[5]{%
  \foreach \i in {0,...,5}{%
    \ifnum\i=5
      \node[circle, draw=#3, line width=1.4pt, fill=white, minimum size=9mm,
            inner sep=0pt] (n#2\i) at (\i*\dx, #1) {};
    \else
      \node[circle, draw=lgry, line width=1pt, fill=white, minimum size=9mm,
            inner sep=0pt] (n#2\i) at (\i*\dx, #1) {};
    \fi
  }
  \foreach \i [evaluate=\i as \j using int(\i+1)] in {0,...,4}{%
    \draw[-{Stealth[length=2.4mm]}, lgry, line width=1pt]
          (n#2\i) to[bend #5=32] (n#2\j);
  }
  \node[#3, font=\footnotesize] at (0.5*\dx, #1+0.72) {#4};
  \foreach \r in {0,1}{%
    \pgfmathsetmacro{\ty}{#1 - \thdrop - \r*\rowsep}
    \foreach \i in {0,...,5}{%
      \ifnum\i=5
        \node[inner sep=0pt, draw=#3, line width=1.4pt]
              at (\i*\dx, \ty) {\includegraphics[width=\tw]{mk_#2_\r_\i}};
      \else
        \node[inner sep=0pt] at (\i*\dx, \ty)
              {\includegraphics[width=\tw]{mk_#2_\r_\i}};
      \fi
    }
  }
}

%% file: figures/teaser/fig_teaser_fancy.tex
\newlength{\tzW}\setlength{\tzW}{2.0cm}
\definecolor{glow}{HTML}{F5B041}\definecolor{keyline}{HTML}{FFFFFF}\definecolor{cone}{HTML}{95A5A6}
\definecolor{slate}{HTML}{2C3E50}\definecolor{accent}{HTML}{C0392B}
\tikzset{
  card/.style={inner sep=0pt, transform shape, draw=keyline, line width=0.6pt,
               drop shadow={opacity=0.35, shadow xshift=1.2pt, shadow yshift=-1.4pt, fill=black}},
  float/.style={inner sep=0pt, transform shape},
  atlascard/.style={inner sep=0pt, draw=keyline, line width=1.0pt, drop shadow={opacity=0.5, shadow xshift=1.5pt, shadow yshift=-2pt, fill=black}},
  atlasfloat/.style={inner sep=0pt},
  deckpersp/.style={yslant=0.16, xslant=-0.22},
  flatdeck/.style={yslant=0.0, xslant=0.0},
  title/.style={font=\small\scshape, text=slate},
  badge/.style={rounded rectangle, fill=slate, text=white, font=\scriptsize\scshape, inner xsep=6pt, inner ysep=2.2pt},
}
\newcommand{\contactshadow}[4]{\fill[black, path fading=circle with fuzzy edge 20 percent, opacity=0.5] (#1,#2) ellipse (#3 and #4);}
\newcommand{\fcard}[7]{\node[#1, rotate=#7] at (#4,#5) {\includegraphics[width=#6\tzW]{figures/teaser/#2_ind#3.png}};}
\newcommand{\fancycolumn}[5][deckpersp]{%
\begin{scope}[shift={(#4,0)}]
  \fill[cone, opacity=0.12] (-1.55,0.55) .. controls (-0.95,1.6) and (-0.8,2.2) .. (-0.78,2.6) -- (0.78,2.6) .. controls (0.8,2.2) and (0.95,1.6) .. (1.55,0.55) -- cycle;
  \begin{scope}[#1]
    \ifthenelse{\equal{#2}{float}}{\contactshadow{0}{-0.75}{1.35}{0.22}}{}
    \fcard{#2}{#3}{05}{-1.05}{0.95}{0.70}{-4}
    \fcard{#2}{#3}{04}{1.05}{0.90}{0.70}{5}
    \fcard{#2}{#3}{03}{-0.55}{0.55}{0.78}{-2}
    \fcard{#2}{#3}{02}{0.60}{0.50}{0.78}{3}
    \fcard{#2}{#3}{01}{0.0}{0.05}{0.90}{0}
  \end{scope}
  \shade[inner color=glow!65, outer color=white, opacity=0.95] (0,3.6) circle (1.75);
  \ifthenelse{\equal{#2}{float}}{\contactshadow{0}{2.62}{1.0}{0.16}\node[atlasfloat] at (0,3.6) {\includegraphics[width=1.2\tzW]{figures/teaser/#3_atlas.png}};}%
                                {\node[atlascard] at (0,3.6) {\includegraphics[width=1.2\tzW]{figures/teaser/#3_atlas.png}};}
  \node[badge] at (0,5.05) {intrinsic atlas};
  \node[title] at (0,-1.35) {#5};
\end{scope}}
\begin{tikzpicture}[x=1cm,y=1cm]
  \fancycolumn[flatdeck]{float}{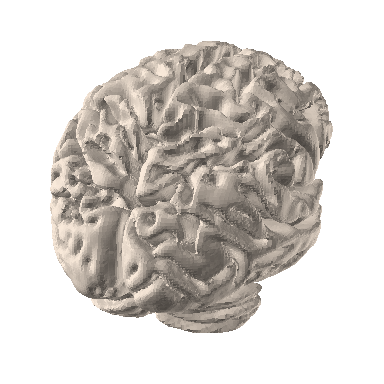}{0}{3D brain MRI}
  \fancycolumn{card}{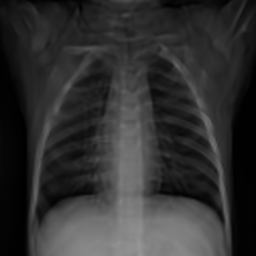}{4.6}{chest X-ray}
  \fancycolumn[flatdeck]{float}{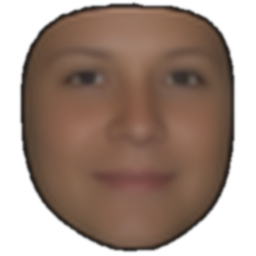}{9.2}{faces}
\end{tikzpicture}

%% file: figures/funnel/fig_funnel.tex
\newcommand{\funnelNS}{50}
\definecolor{seedA}{HTML}{1F77B4}\definecolor{seedB}{HTML}{FF7F0E}\definecolor{seedC}{HTML}{2CA02C}
\definecolor{seedD}{HTML}{D62728}\definecolor{seedE}{HTML}{9467BD}\definecolor{seedF}{HTML}{8C564B}
\pgfplotsset{
  funnel axis/.style={
    width=0.52\linewidth, height=0.42\linewidth,
    y dir=reverse, ymin=-2, ymax=\funnelNS+3,
    xmin=-3.3, xmax=3.3,
    xtick={-3.0043,-2.0043,-1.0414,0,1.0414,2.0043,3.0043},
    xticklabels={$-10^{3}$,,$-10^{1}$,$0$,$10^{1}$,,$10^{3}$},
    ytick={0,10,20,30,40,50},
    grid=both, grid style={black!12}, tick label style={font=\scriptsize},
    label style={font=\scriptsize}, title style={font=\small, align=center, yshift=-1pt},
    every axis plot/.append style={line width=0.95pt, mark size=1.1pt, mark repeat=3, mark phase=1, mark options={solid}, smooth},
    axis line style={black!60},
  },
  seed/.style n args={2}{color=#1, mark=*, mark options={fill=#1, draw=#1}, table/x=u#2, table/y=step},
  init/.style={only marks, mark=o, mark size=3.2pt, line width=1.1pt},
}
\newcommand{\funnelpanel}[2]{%
  \addplot[seed={seedA}{0}] table {figures/funnel/#1_#2.dat};
  \addplot[seed={seedB}{1}] table {figures/funnel/#1_#2.dat};
  \addplot[seed={seedC}{2}] table {figures/funnel/#1_#2.dat};
  \addplot[seed={seedD}{3}] table {figures/funnel/#1_#2.dat};
  \addplot[seed={seedE}{4}] table {figures/funnel/#1_#2.dat};
  \addplot[seed={seedF}{5}] table {figures/funnel/#1_#2.dat};
  \addplot[init, seedA] table[x=u0, y=step, restrict expr to domain={\thisrow{step}}{0:0}] {figures/funnel/#1_#2.dat};
  \addplot[init, seedB] table[x=u1, y=step, restrict expr to domain={\thisrow{step}}{0:0}] {figures/funnel/#1_#2.dat};
  \addplot[init, seedC] table[x=u2, y=step, restrict expr to domain={\thisrow{step}}{0:0}] {figures/funnel/#1_#2.dat};
  \addplot[init, seedD] table[x=u3, y=step, restrict expr to domain={\thisrow{step}}{0:0}] {figures/funnel/#1_#2.dat};
  \addplot[init, seedE] table[x=u4, y=step, restrict expr to domain={\thisrow{step}}{0:0}] {figures/funnel/#1_#2.dat};
  \addplot[init, seedF] table[x=u5, y=step, restrict expr to domain={\thisrow{step}}{0:0}] {figures/funnel/#1_#2.dat};
  \draw[black!40] (axis cs:0,-2) -- (axis cs:0,\funnelNS+3);
}
\begin{tikzpicture}
\begin{groupplot}[group style={group size=2 by 1, horizontal sep=0.6cm, ylabels at=edge left, yticklabels at=edge left}, funnel axis]
\nextgroupplot[title={posterior-mean iteration}, ylabel={\small step}, xlabel={\small deviation from consensus}]
  \funnelpanel{mean}{xt}
  \addplot[only marks, mark=star, mark size=6pt, line width=1.2pt, black] coordinates {(0,\funnelNS)};
\nextgroupplot[title={stochastic sampling}, xlabel={\small deviation from consensus}]
  \funnelpanel{stoch}{xt}
  \addplot[only marks, mark=square*, mark size=2.4pt, seedA] coordinates {(0.118,\funnelNS)};
  \addplot[only marks, mark=square*, mark size=2.4pt, seedB] coordinates {(0.102,\funnelNS)};
  \addplot[only marks, mark=square*, mark size=2.4pt, seedC] coordinates {(-0.351,\funnelNS)};
  \addplot[only marks, mark=square*, mark size=2.4pt, seedD] coordinates {(0.214,\funnelNS)};
  \addplot[only marks, mark=square*, mark size=2.4pt, seedE] coordinates {(0.030,\funnelNS)};
  \addplot[only marks, mark=square*, mark size=2.4pt, seedF] coordinates {(-0.018,\funnelNS)};
\end{groupplot}
\end{tikzpicture}

%% file: figures/fig_algo.tex
\begin{tikzpicture}[font=\normalsize]

\node[font=\large] at (0, 1.55) {$x_T$};
\node[font=\large] at (5*\dx, 1.55) {$x_0$};
\foreach \i in {0,...,5}{%
  \node[lgry, font=\scriptsize] at (\i*\dx, 1.15) {$t=\ts{\i}$};
}

\chainblock{0}{std}{ddpmcol}{$x_{t-1}=\mu_\theta+\sigma_t z$}{left}
\node[ddpmcol, font=\bfseries\small, rotate=90, align=center]
  at (-1.9, {0-\thdrop-0.5*\rowsep})
  {Standard DDPM};
\node[lgry, font=\scriptsize, rotate=90] at (-0.95, {0-\thdrop}) {seed 1};
\node[lgry, font=\scriptsize, rotate=90] at (-0.95, {0-\thdrop-\rowsep}) {seed 4};
\draw[ddpmcol, line width=1.2pt]
  ($(5*\dx+1.05, {0-\thdrop+0.85})$) -- ($(5*\dx+1.05, {0-\thdrop-\rowsep-0.85})$);
\node[ddpmcol, font=\bfseries\small, align=left, anchor=west]
  at (5*\dx+1.2, {0-\thdrop-0.5*\rowsep})
  {2 different\\outputs\\(corr 0.66)};

\draw[lgry!60, dashed, line width=0.8pt]
  (-1.4, {0-\thdrop-\rowsep-1.15}) -- (5*\dx+1.0, {0-\thdrop-\rowsep-1.15});

\pgfmathsetmacro{\yB}{0-\thdrop-\rowsep-2.15}
\chainblock{\yB}{ours}{ourscol}{$x_{t-1}=\mu_\theta$}{left}
\node[ourscol, font=\bfseries\small, rotate=90, align=center]
  at (-1.9, {\yB-\thdrop-0.5*\rowsep})
  {Atlas Construction};
\node[lgry, font=\scriptsize, rotate=90] at (-0.95, {\yB-\thdrop}) {seed 1};
\node[lgry, font=\scriptsize, rotate=90] at (-0.95, {\yB-\thdrop-\rowsep}) {seed 4};
\draw[ourscol, line width=1.2pt]
  ($(5*\dx+1.05, {\yB-\thdrop+0.85})$) -- ($(5*\dx+1.05, {\yB-\thdrop-\rowsep-0.85})$);
\node[ourscol, font=\bfseries\small, align=left, anchor=west]
  at (5*\dx+1.2, {\yB-\thdrop-0.5*\rowsep})
  {same template\\every seed\\(corr 0.9999)};


\end{tikzpicture}

%% file: appendix.tex

\section{Mechanism: Contraction Rate and Controls}
\label{app:mechanism_full}

\paragraph{Contraction rate.}
The posterior-mean update is $x_{t-1}=c_{1,t}\,\hat x_0(x_t)+c_{2,t}\,x_t$ with the DDPM coefficients
$c_{1,t}=\sqrt{\bar\alpha_{t-1}}\,\beta_t/(1-\bar\alpha_t)$ and $c_{2,t}=\sqrt{\alpha_t}\,(1-\bar\alpha_{t-1})/(1-\bar\alpha_t)$, which satisfy
$c_{1,t}+c_{2,t}\sqrt{\bar\alpha_t}=\sqrt{\bar\alpha_{t-1}}$. For a Gaussian population $x_0\sim\mathcal N(\mu,C)$ the noised marginal is
$\mathcal N\big(\sqrt{\bar\alpha_t}\mu,\ \bar\alpha_t C+(1-\bar\alpha_t)I\big)$, and \cref{eq:tweedie} gives the affine denoiser as:
\begin{equation}
\hat x_0(x_t)=\mu+\sqrt{\bar\alpha_t}\,C\big(\bar\alpha_t C+(1-\bar\alpha_t)I\big)^{-1}\big(x_t-\sqrt{\bar\alpha_t}\mu\big).
\end{equation}
Let $e_t=(x_t-\sqrt{\bar\alpha_t}\mu)/\sqrt{\bar\alpha_t}$ be the deviation of the trajectory from the population center in clean-image units. Substituting, along an eigen-direction of $C$ with variance $\lambda$,
\begin{equation}
\label{eq:rho}
e_{t-1}=\rho_t(\lambda)\,e_t,\qquad \rho_t(\lambda)=1-\frac{\beta_t}{\bar\alpha_t\lambda+1-\bar\alpha_t}.
\end{equation}
Since $1-\bar\alpha_t\ge\beta_t$, the factor satisfies $0<\rho_t(\lambda)<1$ at every step and for every $\lambda>0$. Every step therefore brings the state closer to $\mu$, whatever noise it started from, and after the last step the initial distance has been multiplied by $\prod_t\rho_t(\lambda)$. This product is why all seeds end at the same image. The factor also explains three observations.
\begin{itemize}[leftmargin=1.4em,itemsep=1pt,topsep=2pt]
\item \emph{Seeds merge early.} At high noise $\bar\alpha_t\approx0$ and $\rho_t\approx1-\beta_t$ whatever $\lambda$ is. All seeds are pulled together at the same rate, independent of the population, which is the fast merging of \cref{fig:deviation}.
\item \emph{The differences that remain lie where the population varies most.} At low noise $\rho_t\approx1-\beta_t/\lambda$, which is closer to $1$ when $\lambda$ is large, so these directions shrink slowest.
\item \emph{Ordinary sampling does not collapse.} It applies the same shrinking step and then adds fresh noise of variance $\sigma_t^2$, which brings the variance back to that of the marginal, $\bar\alpha_{t-1}\lambda+1-\bar\alpha_{t-1}$. Removing this noise is our only change, and it is what lets the shrinking accumulate.
\end{itemize}
The formula predicts the experiment of \cref{fig:toy} ($T{=}1000$, linear schedule, variances $0.084$ and $0.716$ along the two axes). There $\prod_t\rho_t$ is $3.4{\times}10^{-6}$ and $2.9{\times}10^{-5}$. A seed starts about $1/\sqrt{\bar\alpha_T}=157$ from the center in clean-image units, so the endpoints should lie $4.1{\times}10^{-3}$ from the center on average. We measure $3.8{\times}10^{-3}$. With a shorter schedule ($T{=}400$) the shrinking is incomplete. The predicted leftover spread is $0.095$ along the wide axis and $0.011$ along the narrow one, and the measured spread is $0.079$, elongated along the wide axis as the second point predicts.

\begin{table}[!ht]
\centering\footnotesize
\caption{\textbf{Controls isolating the operator from the network.} (a) The 2D populations of \cref{fig:toy,fig:toy_two}, $300$ seeds: spread of the endpoints around their centroid, and offset of that centroid from the true center (per template for the mixture). (b) Brain-T1 generator: cross-seed NCC over $5$ seeds when noise is re-injected at a fraction $\eta$ of the schedule's standard deviation. (c) 3D shape generators, $10$ seeds: cross-seed NCC of three samplers on the same network.}
\label{tab:controls}
\setlength{\tabcolsep}{4pt}
\begin{subtable}[b]{\textwidth}
\centering
\setlength{\abovecaptionskip}{0em}
\setlength{\belowcaptionskip}{0.3em}
\caption{spread\,/\,offset}
\label{tab:controls_a}
\begin{tabular}{l cc cc}
\toprule
 & \multicolumn{2}{c}{one template (data spread $0.749$)} & \multicolumn{2}{c}{two templates ($0.309$)} \\
\cmidrule(lr){2-3}\cmidrule(lr){4-5}
sampler & learned score & exact score & learned score & exact score \\
\midrule
stochastic            & 0.740\,/\,0.018 & --              & 0.263\,/\,0.041 & -- \\
DDIM ($\eta{=}0$)      & 0.730\,/\,0.032 & --              & 0.299\,/\,0.077 & -- \\
posterior mean         & 0.003\,/\,0.009 & 0.004\,/\,0.0002 & 0.037\,/\,0.170 & 0.036\,/\,0.197 \\
posterior mean, conditioned & -- & --                       & 0.0003\,/\,0.024 & 0.0005\,/\,0.0001 \\
\bottomrule
\end{tabular}
\end{subtable}

\begin{subtable}[t]{.35\textwidth}
\centering
\setlength{\abovecaptionskip}{0em}
\setlength{\belowcaptionskip}{0.3em}
\caption{noise re-injection, brain T1}
\begin{tabular}{l c}
\toprule
$\eta$ & cross-seed NCC \\
\midrule
 1 (DDPM) & 0.468 \\ 
 0.5  & 0.672 \\ 
 0.25 & 0.777 \\ 
 0.1  & 0.921\\ 
 0.05 & 0.968\\
 0 (ours) & 0.9999 \\
\bottomrule
\end{tabular}
\end{subtable}
\begin{subtable}[t]{.62\textwidth}
\centering
\setlength{\abovecaptionskip}{0em}
\setlength{\belowcaptionskip}{0.3em}
\caption{same network, three samplers}
\begin{tabular}{l ccccc}
\toprule
 & \multicolumn{3}{c}{KeypointNet} & \multicolumn{2}{c}{ModelNet} \\
\cmidrule(lr){2-4}\cmidrule(lr){5-6}
sampler & chair & car & airplane & chair & airplane \\
\midrule
stochastic        & 0.51 & 0.87 & 0.74 & 0.35 & 0.76 \\
DDIM ($\eta{=}0$)  & 0.39 & 0.78 & 0.64 & 0.26 & 0.70 \\
posterior mean     & 1.00 & 1.00 & 1.00 & 1.00 & 1.00 \\
\bottomrule
\end{tabular}
\end{subtable}
\end{table}

\paragraph{Several templates.}
The drift of \cref{fig:toy_two} has the same origin. The score of a mixture weights each component by its responsibility, and at high noise the responsibilities are nearly uniform, so every seed is first drawn toward the global mean. Once the components separate, each seed contracts within its own component by \cref{eq:rho}, but the late steps are too small to undo the earlier pull, which leaves each endpoint $0.20$ from its template (\cref{tab:controls_a}).

\paragraph{Controls.}
\Cref{tab:controls} separates the operator from the network in three ways. (a) With the exact score in place of a trained network the collapse is the same, so it is not an artifact of learning. (b) Convergence degrades continuously as noise is put back, so the noise term is what it depends on. (c) On the same trained network DDIM, which is also deterministic, does not converge, so determinism is not the cause.

\begin{figure}[!ht]
\centering
\begin{subfigure}{0.2\linewidth}\includegraphics[width=\linewidth]{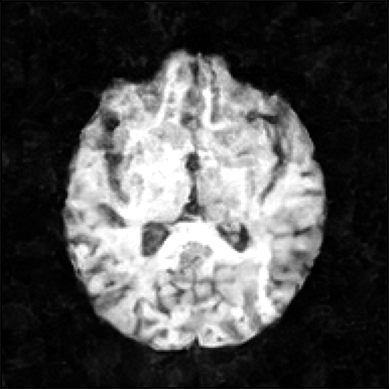}\caption{sample, seed 0}\end{subfigure}
\begin{subfigure}{0.2\linewidth}\includegraphics[width=\linewidth]{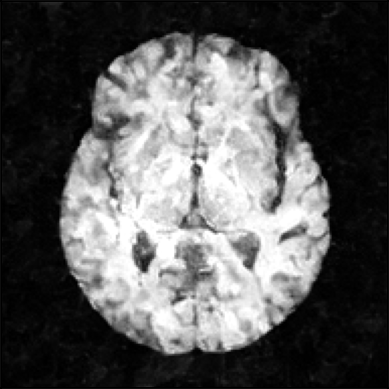}\caption{sample, seed 1}\end{subfigure}
\begin{subfigure}{0.2\linewidth}\includegraphics[width=\linewidth]{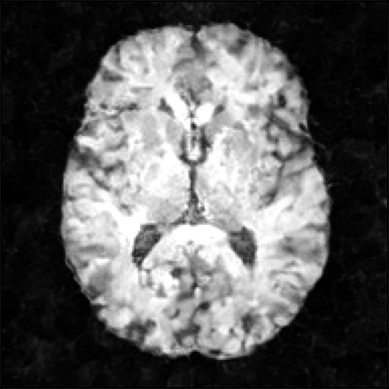}\caption{sample, seed 2}\end{subfigure}\\[2pt]
\begin{subfigure}{0.2\linewidth}\includegraphics[width=\linewidth]{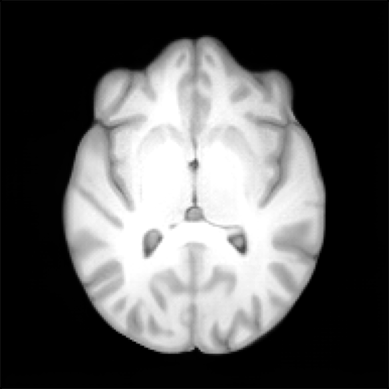}\caption{intrinsic atlas}\end{subfigure}
\begin{subfigure}{0.2\linewidth}\includegraphics[width=\linewidth]{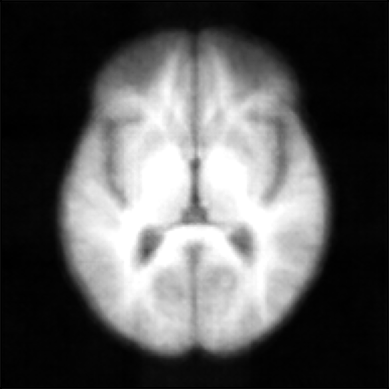}\caption{mean, $50$ samples}\end{subfigure}
\begin{subfigure}{0.2\linewidth}\includegraphics[width=\linewidth]{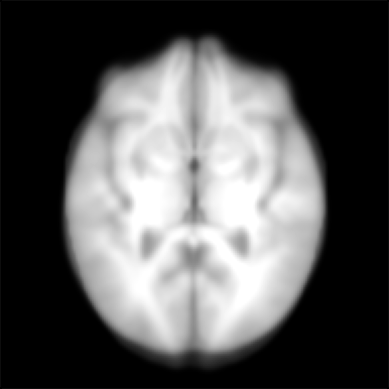}\caption{training mean}\end{subfigure}
\caption{\textbf{Samples, atlas, and averages from one generator.}}
\label{fig:s1}
\end{figure}

\paragraph{An instance, not an average.}
\Cref{fig:s1} contrasts the three objects obtainable from the same brain-T1 generator. Stochastic samples are distinct individuals (pairwise NCC $0.47$ over $50$ seeds). Their voxelwise mean is a blur. The intrinsic atlas correlates with that mean (NCC $0.84$) but is about $8\times$ sharper (variance of the Laplacian within the brain: atlas $0.020$, mean of $50$ samples $0.0024$, one sample $0.045$).

\section{Evaluation Protocol and Metrics}
\label{app:metrics}

\subsection{Data, generators, and baselines}
\label{app:benchmarks}
In every domain the generator and all constructed baselines see the same training images, and evaluation uses held-out subjects. All atlases are read out at $t^*{=}99$ (\cref{app:tstar}).

\textbf{Brain T1.} Two generators: the pretrained 3D latent diffusion model~\citep{khader2022medical} as released (``off-the-shelf''), and the same model fine-tuned for age-conditioned synthesis on $899$ IXI~\citep{ixi} and OASIS-1~\citep{oasis1} volumes (\cref{app:age_protocol}), read out with its null age token. Neither saw the test cohorts: IBSR18~\citep{ibsr} ($n{=}18$), Mindboggle101~\citep{mindboggle101} ($n{=}80$, excluding its OASIS subjects), and $163$ adult controls of ABIDE-I~\citep{di2014autism}. Baselines built on the $899$ volumes: voxel mean; ANTs template construction ($60$-subject random subset, $4$ iterations, initialized at the voxel mean); the VoxelMorph learned template~\citep{Dalca2019learning}; Aladdin~\citep{ding2022aladdin} ($100$ epochs). MultiMorph~\citep{abulnaga2025multimorph} is evaluated through its published atlas.

\textbf{Faces.} A 2D DDPM ($128^2$, $40$k steps, batch $48$) trained on $30{,}000$ CelebA~\citep{liu2015faceattributes} training faces cropped to the face oval; test set, $1{,}000$ CelebA test faces with their $5$ landmarks. Baselines on the same images: pixel mean, ANTs template construction, VoxelMorph template ($20$k steps), Aladdin ($150$ epochs).

\textbf{Chest X-ray.} A 2D DDPM ($128^2$, $30$k steps, batch $32$) trained on $1{,}341$ normal pediatric films~\citep{kermany2018identifying}; test set is the $138$ Montgomery~\citep{jaeger2014two}, with $8$ landmarks derived from their lung masks (apex, costophrenic angle, and lateral and medial extremes of each lung).

\textbf{3D shapes.} Voxel DDPMs ($32^3$ occupancy) trained per category on the KeypointNet~\citep{you2020keypointnet} training split ($799$ chairs, $801$ cars); test sets, $200$ and $201$ shapes with their semantic keypoints. Baselines: voxel mean, VoxelMorph template ($10$k steps), Aladdin ($150$ epochs), and the category templates released by DIF-Net~\citep{deng2021deformed} and DIT~\citep{zheng2021deep}, evaluated under our registration. Part labels come from ShapeNet-Part~\citep{yi2016scalable}. They are transferred to the KeypointNet clouds after aligning the two releases of each model by the best of the $24$ axis-aligned rotations under chamfer distance, and shapes whose alignment is poor or ambiguous are discarded, leaving $254$ chairs and $105$ cars.

\subsection{Registration and metrics}
The presets of \cref{tab:pck_ants} are run with ANTs defaults: \texttt{antsRegistrationSyNQuick[b]} (rigid, affine, B-spline SyN; mutual information) on brain volumes, \texttt{SyNCC} on $128^2$ images, and \texttt{SyN} on occupancy volumes upsampled $\times2$.
Let $u_i$ be the displacement field (mm) of subject $i$'s registration, defined on the template grid, $\phi_i=\mathrm{id}+u_i$, and $M$ the template's foreground mask.

\subsubsection{Metrics for T1 Brain MRI}

\textbf{Dice}. With $S_i^{(k)}$ region $k$ of subject $i$ warped into template space,
\begin{equation}
\mathrm{Dice}=\frac{2}{N(N-1)}\sum_{i<j}\frac1K\sum_{k=1}^K\frac{2\,|S_i^{(k)}\cap S_j^{(k)}|}{|S_i^{(k)}|+|S_j^{(k)}|},
\end{equation}
the pairwise inter-subject convention for atlases that carry no labels of their own~\citep{dey2021generative,abulnaga2025multimorph}. Regions: IBSR18, three tissue classes; Mindboggle101, the $62$ DKT parcels merged into $7$ lobes; ABIDE-I, FreeSurfer \texttt{aparc+aseg} merged into the same $7$ lobes plus $8$ subcortical structures.

\textbf{Centrality, mean displacement, regularity}.
\begin{gather}
\|\bar u\|=\frac1{|M|}\sum_{x\in M}\Big|\frac1N\sum_i u_i(x)\Big|,\qquad
|u|=\frac1{N|M|}\sum_i\sum_{x\in M}|u_i(x)|,\\
\mathrm{SDlogJ}=\frac1N\sum_i\operatorname*{std}_{x\in M}\,\log\!\big(\det\nabla\phi_i(x)+3\big).
\end{gather}
Centrality is the magnitude of the mean field. It vanishes for a template at the center of the cohort, where opposing deformations cancel~\citep{Dalca2019learning,abulnaga2025multimorph}. $|u|$ is the mean magnitude, which never cancels. SDlogJ is the Learn2Reg scorer~\citep{hering2022learn2reg}, including its $+3$ offset, central differences, and two-voxel border crop. Our absolute centrality values are not comparable with those of \citet{abulnaga2025multimorph}, whose atlases are built for the evaluated group. Comparisons within \cref{tab:pck_ants} are valid because every template is scored by the same code on the same cohort.

\subsubsection{Metrics for Other Modalities}
\textbf{PCK}. Landmarks are transferred from one test subject to another through the template, and PCK@$\alpha$ is the fraction landing within $\alpha s$ of the target's annotation. For images, all ordered pairs are used and $s$ is the image side~\citep{peebles2022gan}. For shapes, we follow the 5-shot protocol of DIF-Net~\citep{deng2021deformed}: the keypoints of $5$ fixed source shapes are mapped into template space and averaged per semantic label, then mapped to each test shape, with $s$ the radius of the target's bounding sphere.

\textbf{Label IoU}~\citep{deng2021deformed}. The labeled points of $5$ fixed source shapes and the points of each target are mapped into template space, and each target point takes the majority label of its $10$ nearest source points. Per shape, IoU is averaged over the parts present; we report the mean and median over shapes.

\textbf{Cross-seed consensus.} Pairwise SSIM (NCC where stated) between atlases from independent seeds, on the foreground.

\subsection{Verification and repeated runs}
\label{app:metric_impl}
All brain metrics come from one script and one registration per subject. SDlogJ and the Jacobian determinant are line-for-line ports of the Learn2Reg scorer (\texttt{MDL-UzL/L2R}, commit \texttt{88475095}), and Dice of VoxelMorph's (\texttt{voxelmorph/py/utils.py}, commit \texttt{c4155e1b}); ANTs fields are first converted from physical millimetres to the voxel index frame. A suite of $41$ checks verifies that the ports are syntactically identical to the pinned upstream code and return identical arrays, that the Jacobian agrees with ANTs' own ($r{=}0.99997$), and analytic cases: identity and uniform-scale fields, a hand-computed Dice, and, for centrality, that opposing fields cancel to $\|\bar u\|{=}0$ while $|u|$ does not. Centrality and $|u|$ have no public reference implementation and follow the definitions of \citet{Dalca2019learning} and \citet{abulnaga2025multimorph}.

\begin{table}[!ht]
\centering\footnotesize
\caption{\textbf{Repeated runs behind \cref{tab:pck_ants}.} Range over runs of the headline metric (Dice, PCK@0.1, mean IoU), smallest to largest across templates.}
\label{tab:repeats}
\begin{tabular}{lccc}
\toprule
Domain & test subjects & runs per cell & range over runs \\
\midrule
Brain T1 & 18 / 80 / 163 & 3 & $\le0.0003$ \\
Faces & 1000 & 5--6 & 0.005 \\
Chest X-ray & 138 & 4 & 0.01--0.09 \\
Shapes, PCK (chair / car) & 200 / 201 & 6--9 & 0.02--0.16 / 0.01--0.04 \\
Shapes, label IoU (chair / car) & 254 / 105 & 6 / 3 & 0.02--0.09 / 0.001--0.02 \\
\bottomrule
\end{tabular}
\end{table}

ANTs samples its similarity metric stochastically, so every cell of \cref{tab:pck_ants} is a mean over repeated runs (\cref{tab:repeats}); the $5$ source shapes are fixed across runs and templates. Run-to-run variation is negligible for brain MRI and faces. It is large on the $138$-film chest X-ray set, where only Aladdin's lead exceeds it, and on chairs, where the intrinsic atlas and DIF-Net are separated from every other template but not from each other.

\begin{figure}[!ht]
\centering
\includegraphics[width=\linewidth]{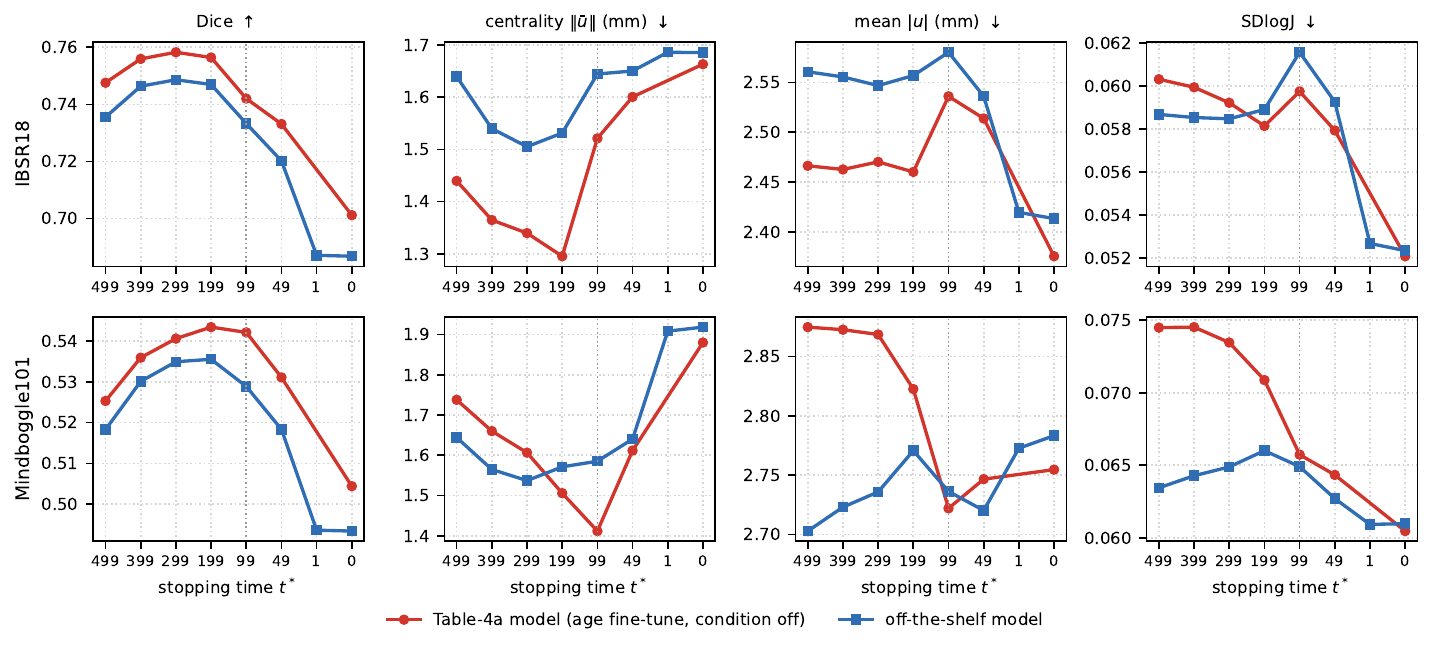}
\caption{\textbf{Stopping-time ablation}. On IBSR18 (top) and Mindboggle101 (bottom), for the fine-tuned (red) and off-the-shelf (blue) generators. Dotted line, the operating point $t^*{=}99$.}
\label{fig:si_tstar}
\end{figure}

\section{Further Results on Brain MRI}

\subsection{Stopping time}
\label{app:tstar}
\begin{figure}[!ht]
\centering
\begin{subfigure}{0.23\linewidth}\includegraphics[width=\linewidth]{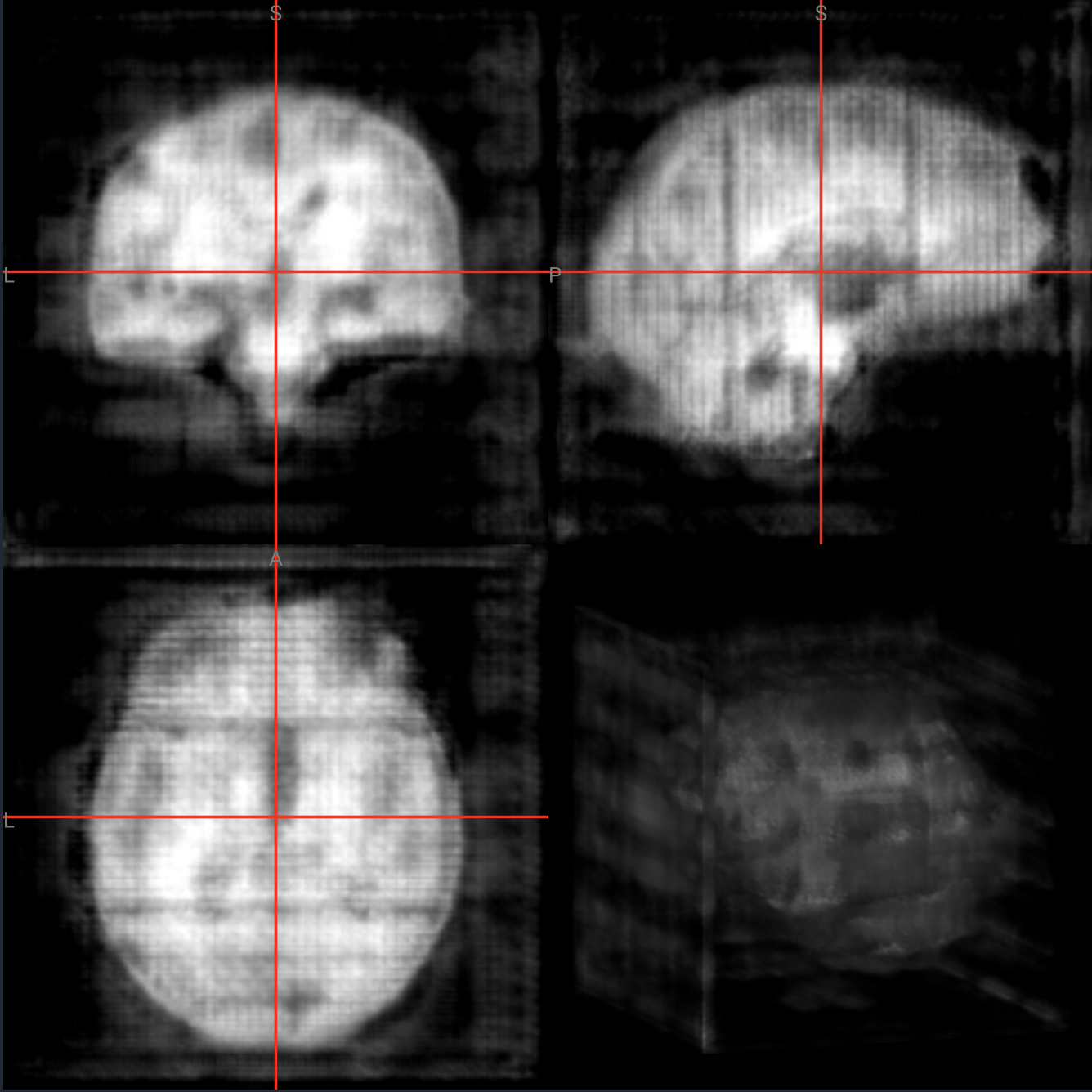}\caption{$t^*{=}799$}\end{subfigure}
\begin{subfigure}{0.23\linewidth}\includegraphics[width=\linewidth]{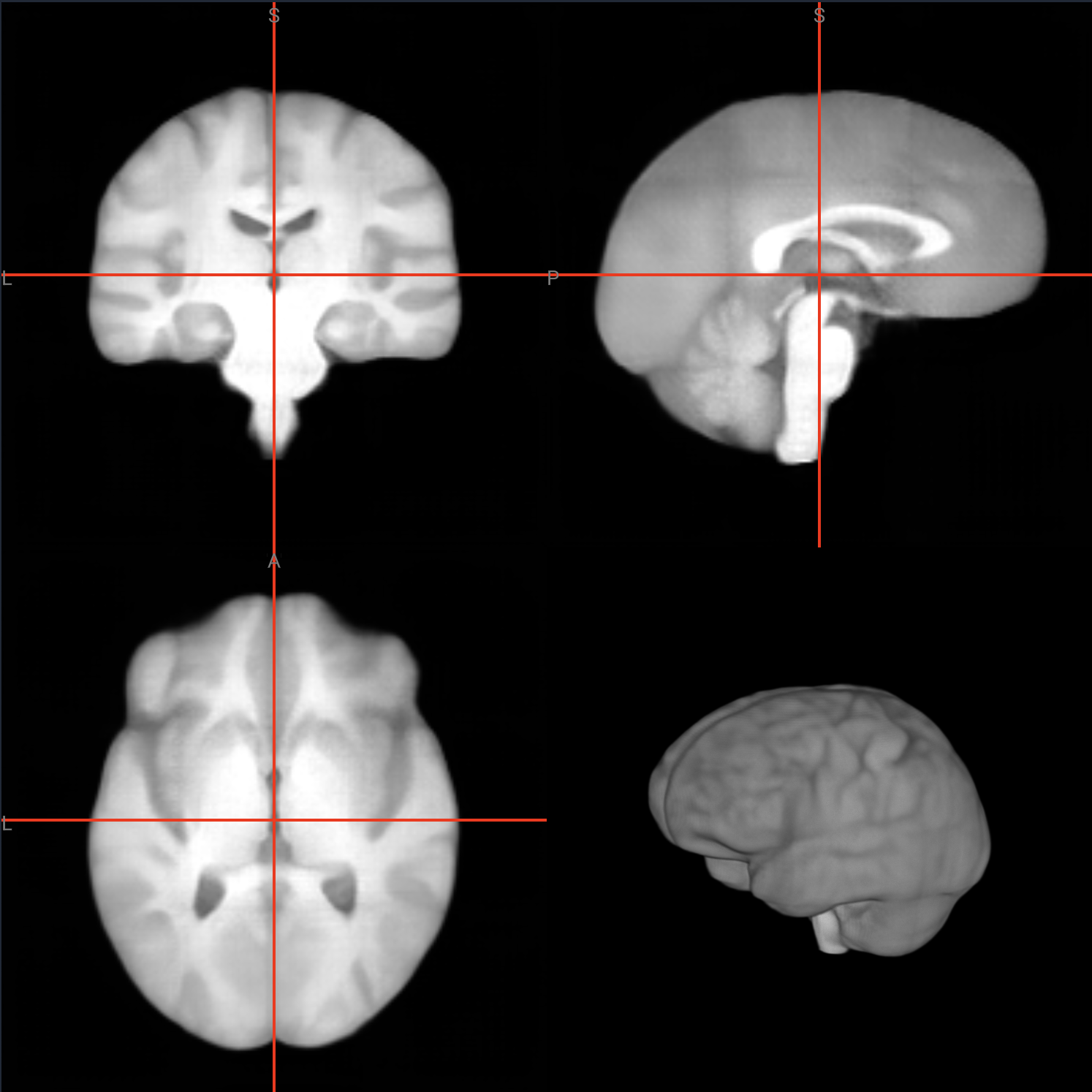}\caption{$499$}\end{subfigure}
\begin{subfigure}{0.23\linewidth}\includegraphics[width=\linewidth]{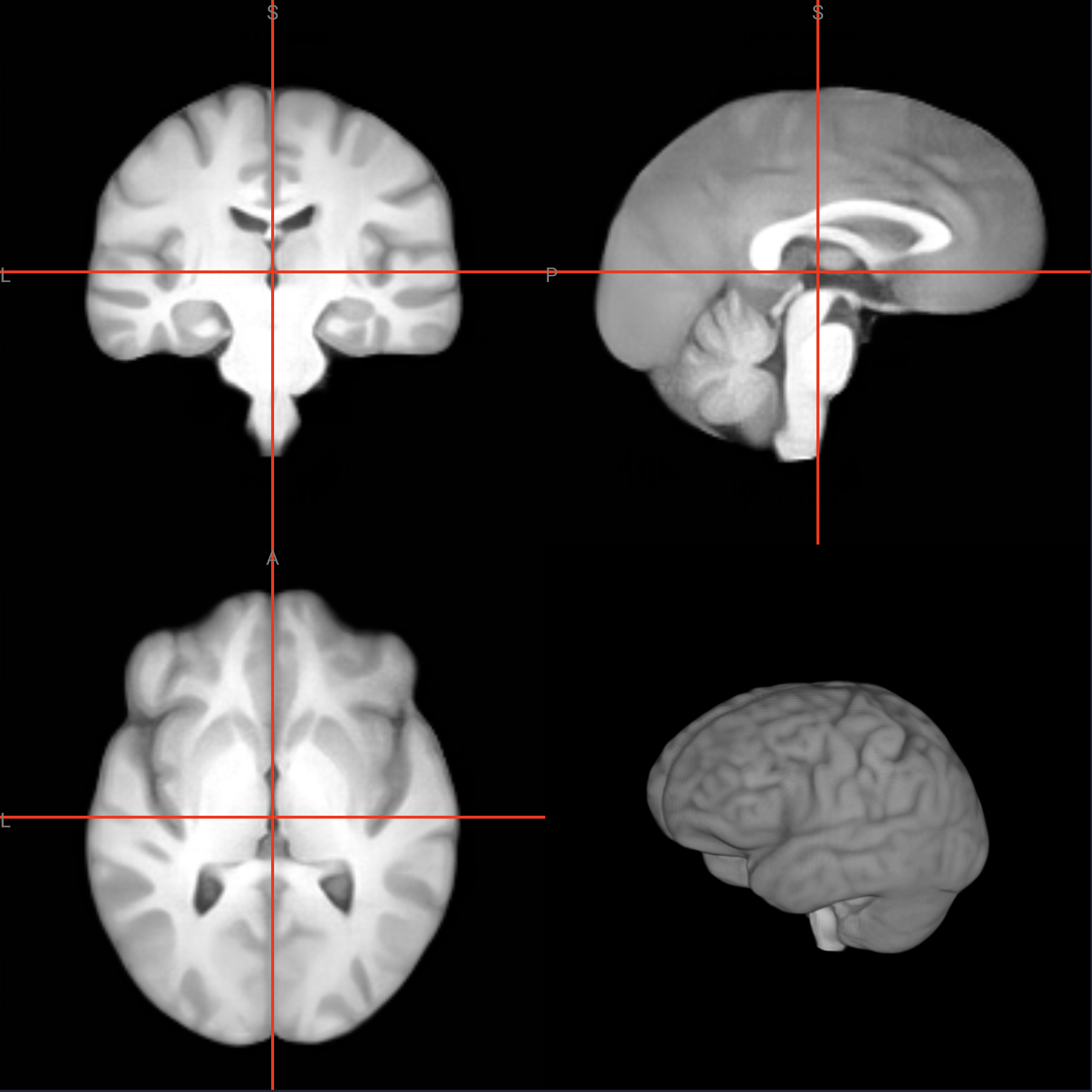}\caption{$299$}\end{subfigure}

\begin{subfigure}{0.23\linewidth}\includegraphics[width=\linewidth]{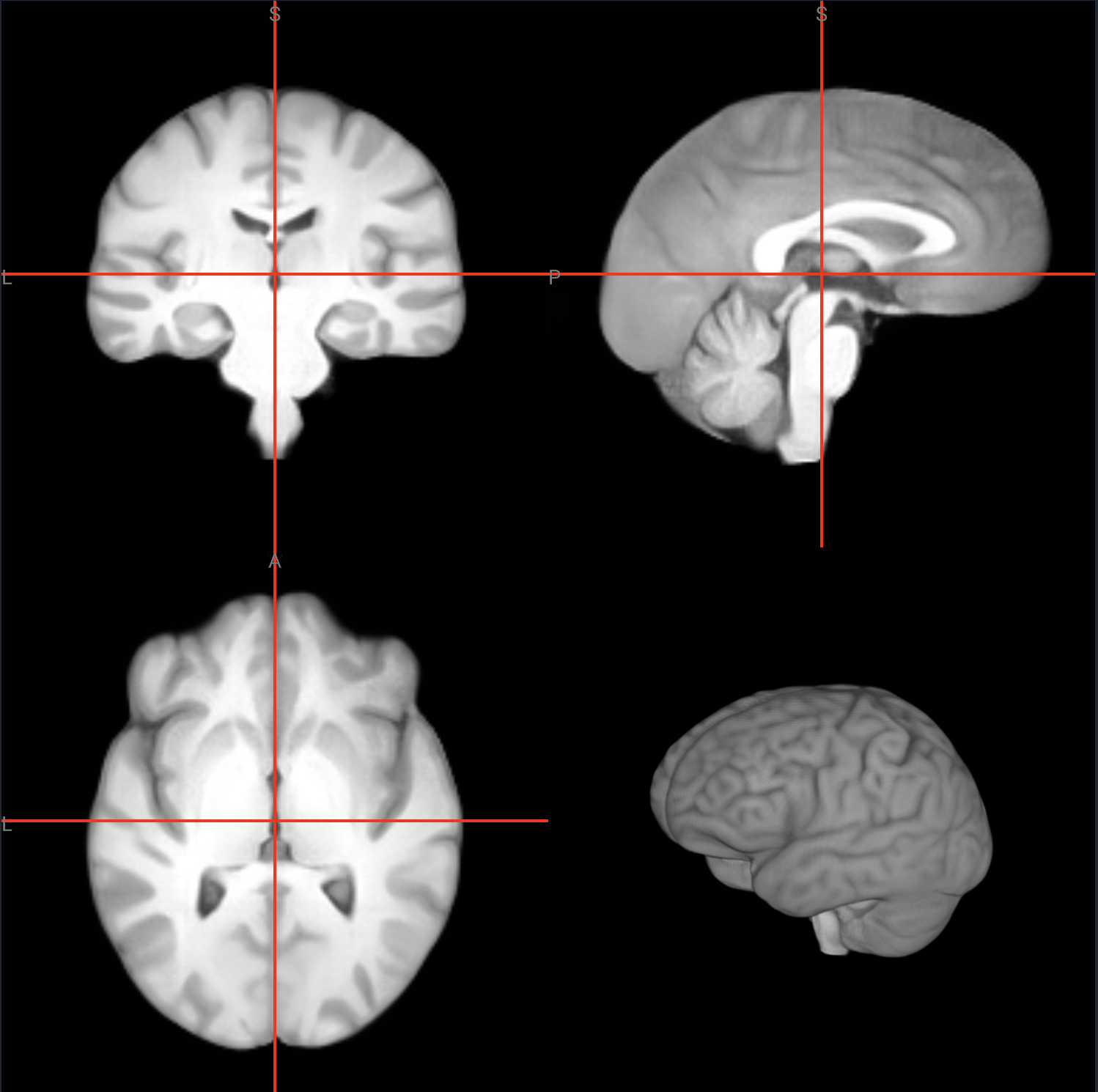}\caption{$199$}\end{subfigure}
\begin{subfigure}{0.23\linewidth}\includegraphics[width=\linewidth]{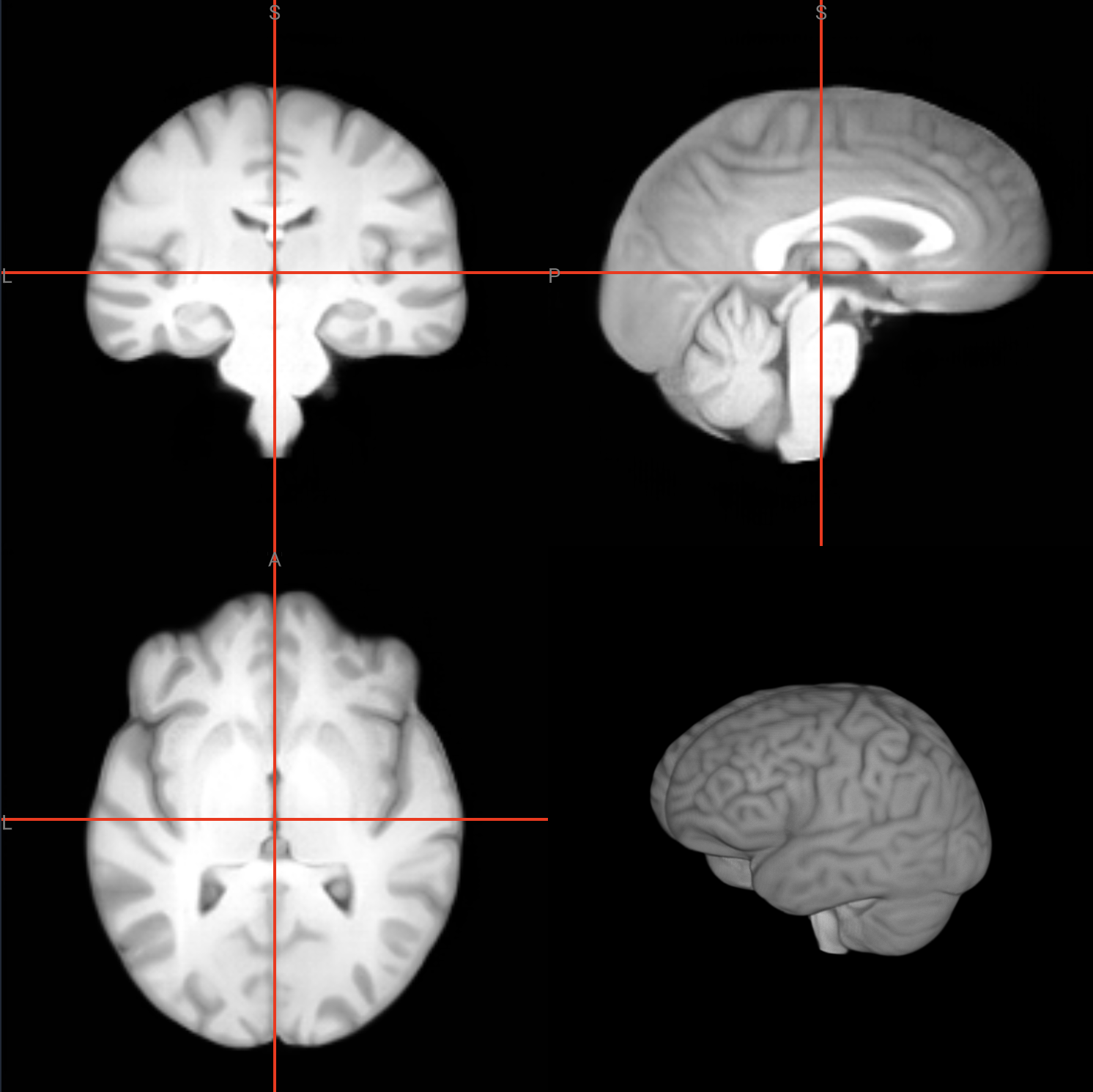}\caption{$99$}\end{subfigure}
\begin{subfigure}{0.23\linewidth}\includegraphics[width=\linewidth]{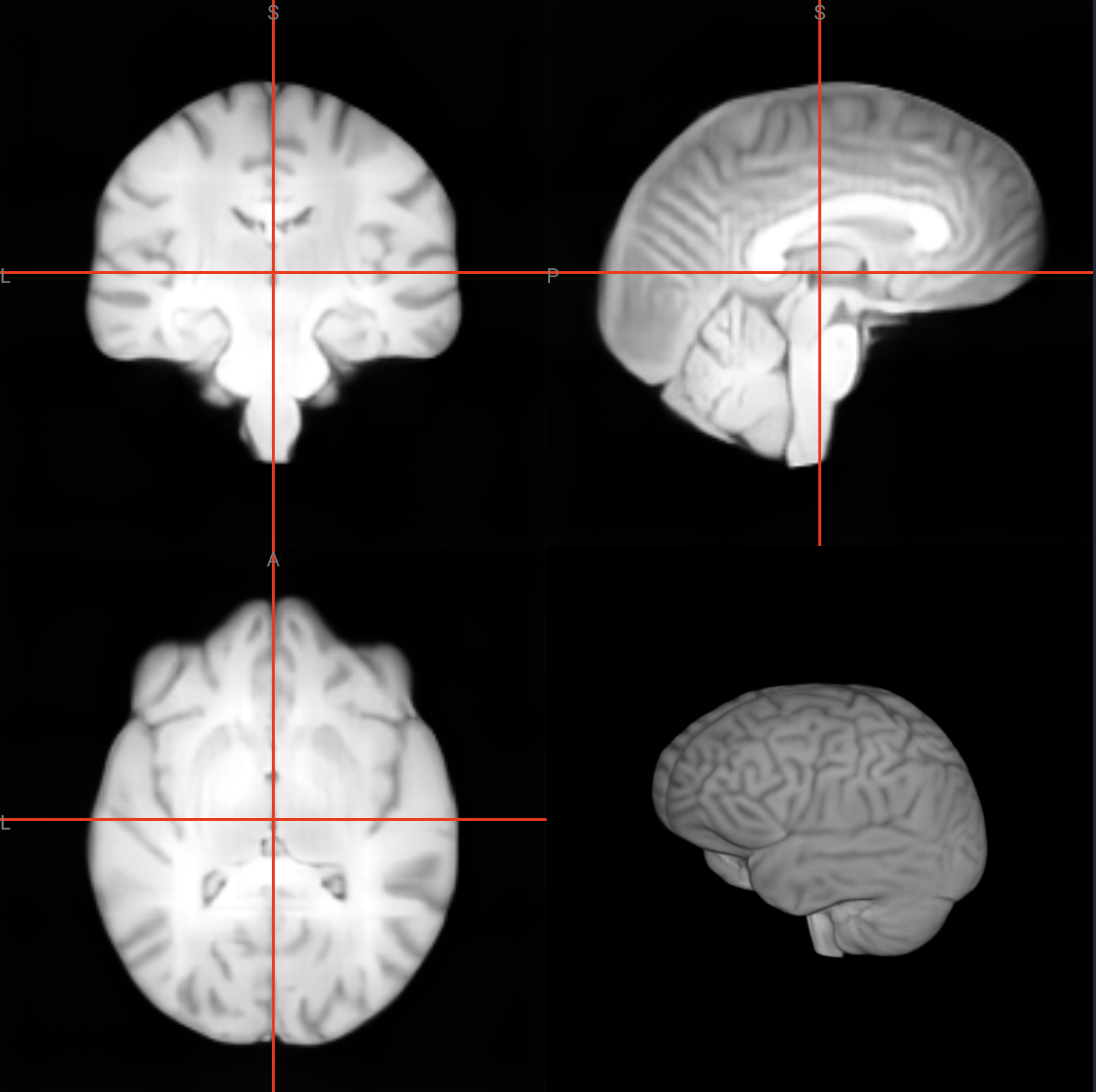}\caption{$0$}\end{subfigure}
\caption{\textbf{The atlas at successive endpoints.} Early endpoints carry coarse population structure, and later ones add anatomical detail.}
\label{fig:main_abl}
\end{figure}

\Cref{fig:si_tstar} repeats the evaluation of \cref{tab:pck_ants} for atlases read out at $t^*\in\{499,399,299,199,99,49,0\}$, for both generators. Dice is flat on $t^*\in[99,499]$ (within $0.016$ on IBSR18 and $0.02$ on Mindboggle101), peaks at $t^*\in[199,299]$, and falls sharply below $t^*{=}49$ ($0.70$ and $0.50$ at $t^*{=}0$). There, SDlogJ and $|u|$ fall together with Dice, the signature of under-registration: the atlas is dominated by subject-specific detail that the engine cannot match, so it deforms less, not better. We use $t^*{=}99$ throughout, a choice made before these runs and not the Dice maximum. Every ranking in \cref{tab:pck_ants} is unchanged anywhere on the plateau. \Cref{fig:main_abl} shows the atlas at successive endpoints.

\subsection{Qualitative comparison}
\label{app:qualitative}
\begin{figure}[!ht]
\centering
\begin{subfigure}{0.246\linewidth}\includegraphics[width=\linewidth]{appendix_figures/abl_mri/MRI_999_99.png}\caption{Ours}\end{subfigure}
\begin{subfigure}{0.249\linewidth}\includegraphics[width=\linewidth,trim={0cm .2cm 0cm .5cm},clip]{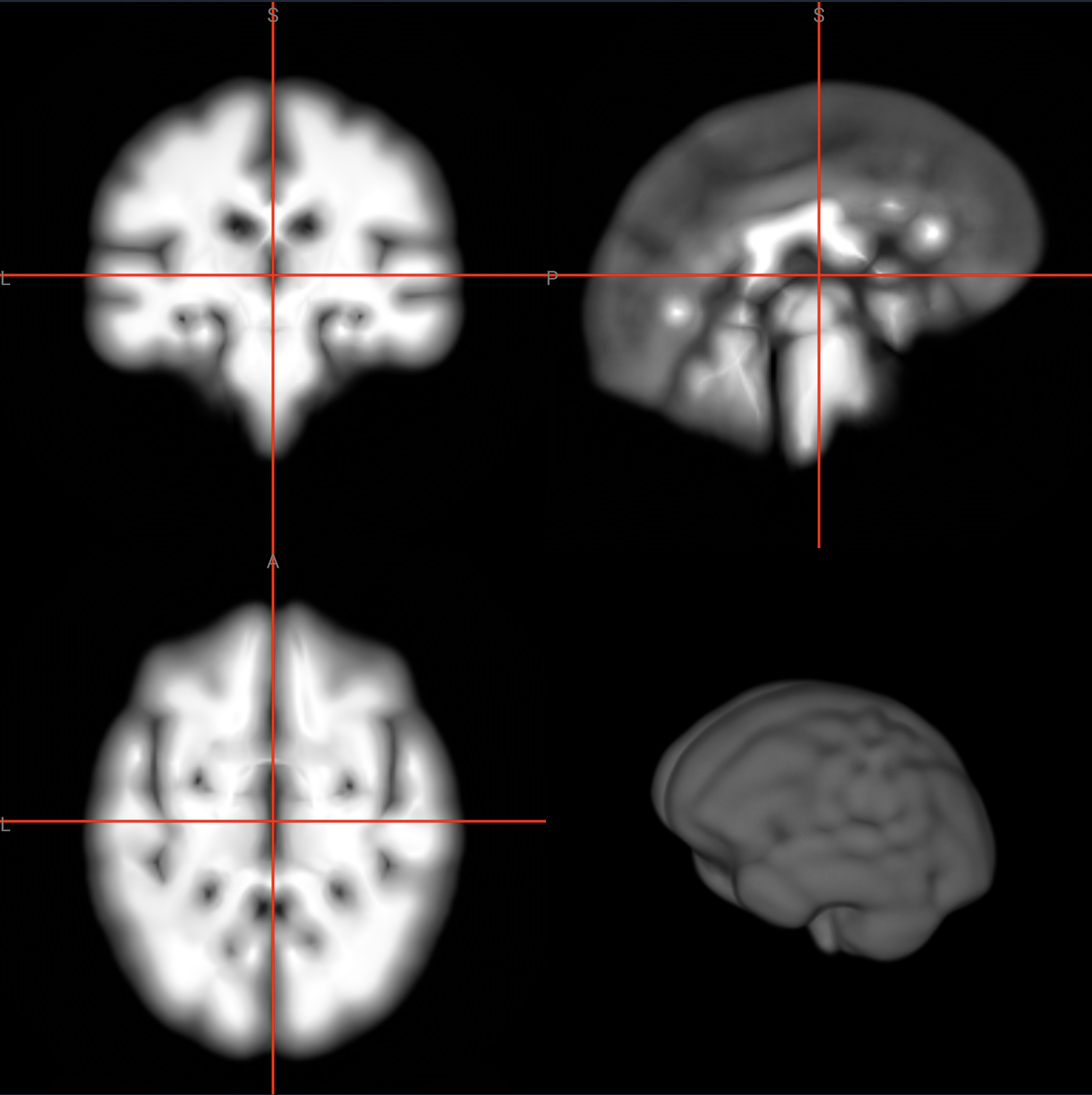}\caption{Aladdin}\end{subfigure}
\begin{subfigure}{0.23\linewidth}\includegraphics[width=\linewidth]{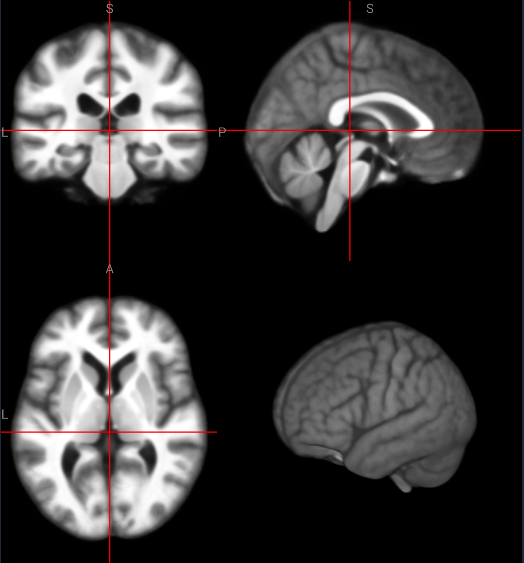}\caption{MultiMorph}\end{subfigure}
\begin{subfigure}{0.23\linewidth}\includegraphics[width=\linewidth,trim={0cm .1cm 0cm 0},clip]{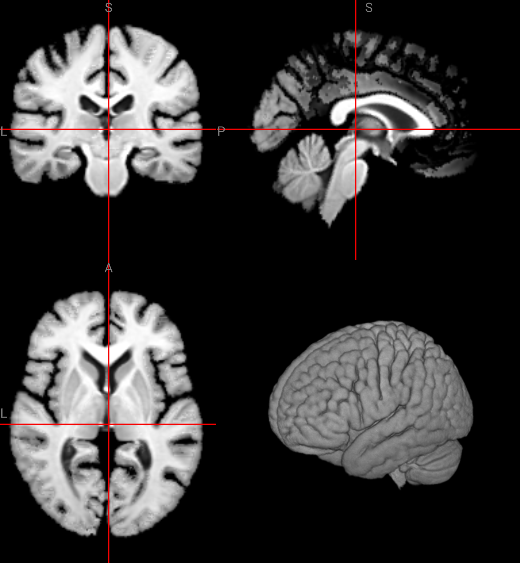}\caption{VoxelMorph}\end{subfigure}
\caption{\textbf{The intrinsic atlas beside established and learned brain atlases.} A clinician judged our atlas anatomically plausible, with no missing or duplicated structures.}
\label{fig:qual}
\end{figure}

\Cref{fig:qual} places the intrinsic atlas beside three comparison templates of \cref{tab:pck_ants}.

\section{Conditioned Atlases}

\subsection{Age-Conditioned Family}
\label{app:age_protocol}

\paragraph{Fine-tuning.} Age is normalized to $[0,1]$, passed through a sinusoidal embedding and a two-layer MLP, added to the time-step embedding, and injected into every residual block through a zero-initialized FiLM layer, so the pretrained model is unchanged at the start. A learned null token implements classifier-free dropout (rate $0.05$). The $899$ volumes (IXI $563$, OASIS-1 $336$) are skull-stripped, affinely registered into the model space, and encoded by the frozen autoencoder. Integer-coded ages receive uniform one-year jitter, and sampling is age-balanced (inverse decade frequency, capped at $3\times$). We train $80$k steps with Adam, learning rate $10^{-5}$ for the pretrained body and $10^{-3}$ for the age modules, with the base model's $L_1$ objective. Recovery degenerates at the edges of the labeled range ($20$ and, more mildly, $80$), so we report ages $22$--$80$ and register to atlases at $25$--$75$.

\paragraph{Registration protocol.} We sample $18$ subjects per decade ($126$ subjects, ages $20$--$89$) and register each to four targets: the atlas at the midpoint of its decade (capped at $75$), a distant-age atlas ($22$ for subjects aged $50$ or older, $75$ otherwise), the unconditioned atlas of the same checkpoint (null age token), and the VoxelMorph template of \cref{tab:pck_ants}. \Cref{tab:age_cohort} splits the cohort by source, because the two sources are in different spaces. IXI scans are in native space, whereas we used OASIS-1's atlas-registered images, which are already warped to Talairach space. The training population therefore has two centers, with IXI the majority ($563$ of $899$), and the intrinsic atlas settles at the majority center, as \cref{sec:discussion} describes for such populations. On the $81$ IXI subjects every intrinsic atlas, age-matched or not, is more central, closer, and more regular than every baseline. On the $45$ OASIS-1 subjects every template, whatever the method, needs more deformation than on IXI. The templates that commit most sharply to the native-space center are the farthest, ours at $5.6$\,mm and MultiMorph, whose published atlas never saw Talairach-space images, at $5.7$\,mm. The voxel mean, the plain average of both sources, is the closest at $3.9$\,mm. This is a mismatch of image space, not a loss of atlas quality. An atlas for Talairach-space subjects calls for a generator conditioned on the source, or trained in that space, in the same way that age conditioning separates the age groups. \Cref{fig:age_utility} therefore reports the IXI subjects, restricted to ages $20$--$79$ ($n{=}78$), with paired Wilcoxon tests.

\begin{table}[!ht]
\centering\footnotesize
\caption{\textbf{Age cohort by source} (two-run means over all seven decades). Dice is tissue overlap from automated segmentation~\citep{Tustison2021}, which fails on the OASIS-1 volumes and is omitted there.}
\label{tab:age_cohort}
\setlength{\tabcolsep}{5pt}
\begin{tabular}{l cccc ccc}
\toprule
 & \multicolumn{4}{c}{IXI ($n{=}81$, native space)} & \multicolumn{3}{c}{OASIS-1 ($n{=}45$, Talairach space)} \\
\cmidrule(lr){2-5}\cmidrule(lr){6-8}
Template & Dice$\uparrow$ & $\|\bar u\|\downarrow$ & $|u|\downarrow$ & SDlogJ$\downarrow$ & $\|\bar u\|\downarrow$ & $|u|\downarrow$ & SDlogJ$\downarrow$ \\
\midrule
voxel mean  & 0.450 & 2.58 & 3.33 & 0.082 & \best{3.58} & \best{3.94} & \best{0.078} \\
ANTs        & 0.453 & 2.74 & 3.43 & 0.068 & 3.96 & 4.35 & 0.085 \\
VoxelMorph  & 0.456 & 2.16 & 2.92 & 0.068 & 3.96 & 4.36 & 0.091 \\
Aladdin     & 0.455 & 2.63 & 3.29 & 0.076 & 3.97 & 4.32 & 0.085 \\
MultiMorph  & \best{0.467} & 1.99 & 2.89 & 0.076 & 5.32 & 5.72 & 0.096 \\
\midrule
Ours, distant age   & 0.455 & 1.54 & 2.53 & 0.058 & 5.31 & 5.62 & 0.081 \\
Ours, unconditioned & 0.456 & \best{1.47} & 2.48 & 0.058 & 5.23 & 5.59 & 0.081 \\
Ours, age-matched   & 0.456 & 1.50 & \best{2.44} & \best{0.057} & 5.34 & 5.66 & 0.080 \\
\bottomrule
\end{tabular}
\end{table}

\subsection{Contrast-conditioned Family}
\label{app:crossmodal}
\begin{figure}[!ht]
\centering
\begin{subfigure}[b]{.3\textwidth}
    \includegraphics[width=\linewidth]{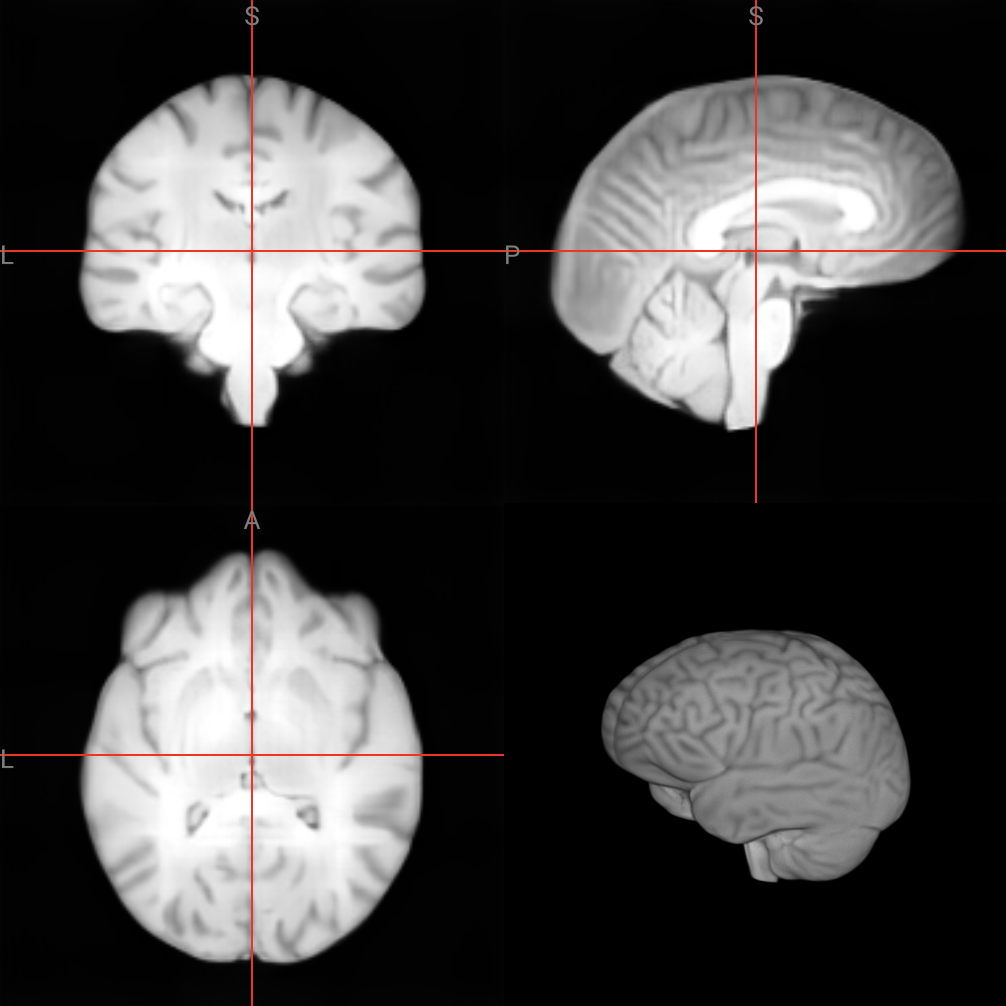}
    \caption{T1-weighted}
\end{subfigure}
\begin{subfigure}[b]{.3\textwidth}
    \includegraphics[width=\linewidth]{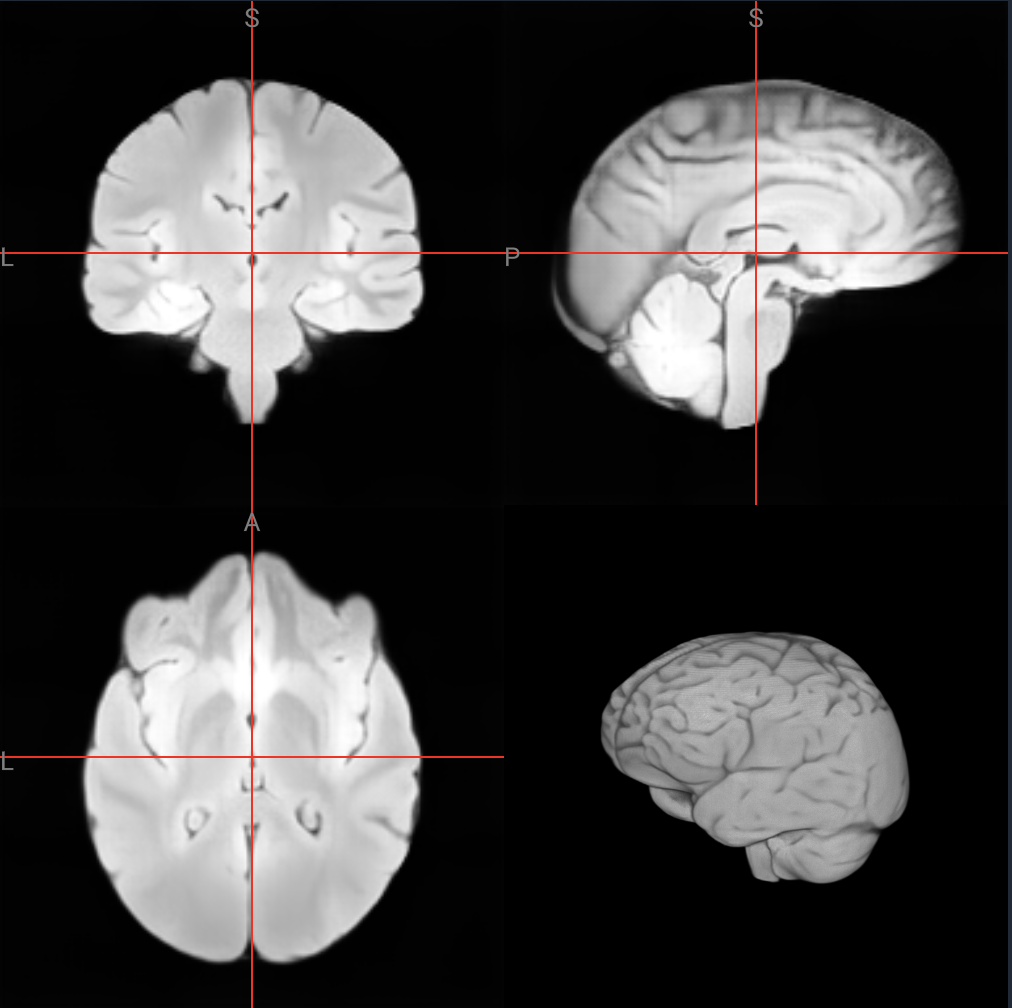}
    \caption{T2-weighted}
\end{subfigure}
\caption{\textbf{Contrast conditioned atlases in one model.}}
\label{fig:t1t2}
\end{figure}

The pretrained 3D model is conditioned on an anatomy class, and T1-weighted and T2-weighted brain MRI are two of its classes. Switching the class token yields a separate atlas for each contrast from the same weights (\cref{fig:t1t2}), each identical across $20$ seeds (cross-seed SSIM $0.9997$ and $0.9992$).

\section{Occupancy as consensus confidence}
\label{app:occupancy}
\begin{figure}[!ht]
\centering
\includegraphics[width=\linewidth]{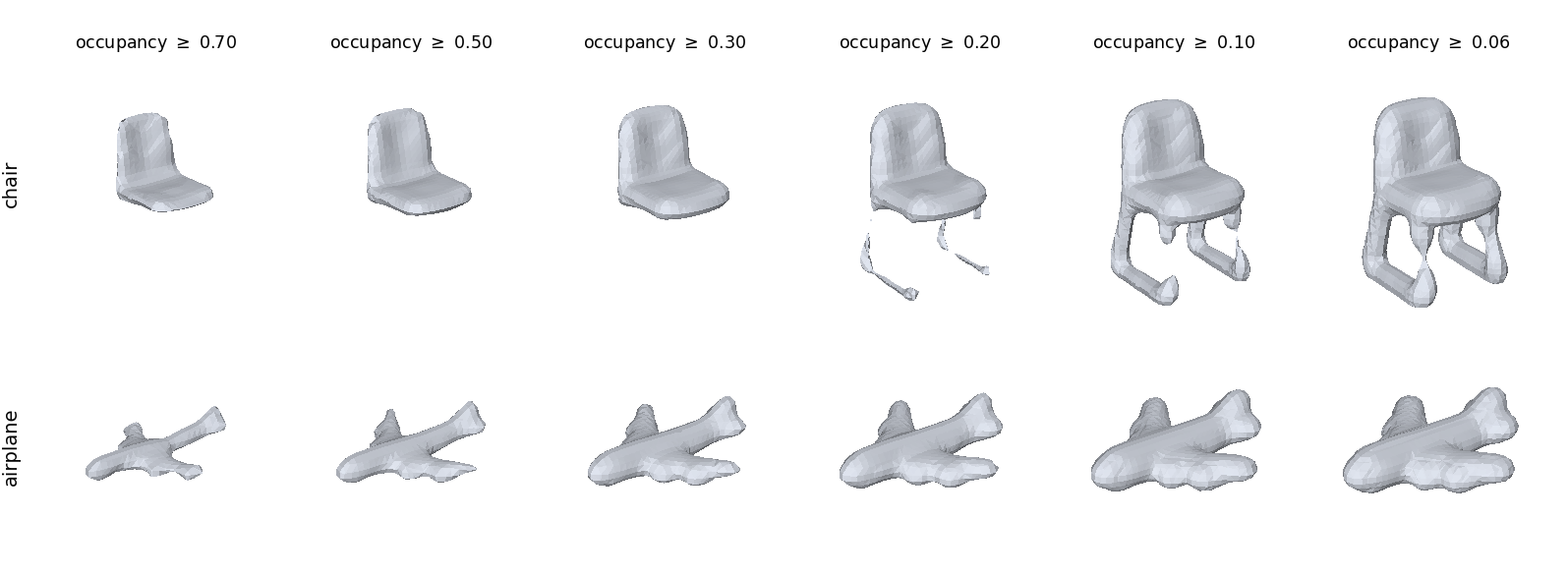}
\caption{\textbf{Atlas isosurfaces at decreasing occupancy.} The chair (top) reveals its legs only below $0.2$, and the airplane (bottom) does not change.}
\label{fig:occupancy_sweep}
\end{figure}

A 3D atlas is a soft occupancy volume, and its values grade how consistently the population shares each structure (\cref{fig:occupancy_sweep}). The legs missing from the chair of \cref{fig:generality} are present in the volume, but only below occupancy $0.2$, whereas the airplane does not change with the level. \Cref{fig:generality} displays the level at which the atlas occupies as many voxels as a median training shape.

\section{Reproducibility}
\label{app:repro}
The 3D medical generator is the pretrained class-conditional latent diffusion model~\citep{khader2022medical}: a U-shaped denoiser (base width $72$, multipliers $[1,1,2,4,8]$, sparse linear attention at the two deepest stages) over the $8$-channel latent space of a $4\times$ vector-quantized autoencoder, conditioned on anatomy class and resolution, with $T{=}1000$. The endpoint is decoded through the frozen autoencoder. One atlas at $32{\times}64{\times}64$ latent resolution takes about two minutes on an A100. The 2D and 3D generators of \cref{app:benchmarks} are pixel- and voxel-space DDPMs with U-Net denoisers, trained from scratch. Code, configurations, and the metric test suite will be released upon publication.